\documentclass[lettersize,journal]{IEEEtran}
\usepackage{amsmath,amssymb,amsfonts}
\usepackage{algorithmic}
\usepackage{algorithm}
\usepackage{array}
\usepackage[caption=false,font=normalsize,labelfont=sf,textfont=sf]{subfig}
\usepackage{textcomp}
\usepackage{stfloats}
\usepackage{url}
\usepackage{verbatim}
\usepackage{graphicx}
\usepackage{orcidlink}
\usepackage{cite}
\usepackage{multirow}
\usepackage{booktabs}
\begin{document}
\title{Robust Real-Time Monocular Human Pose Estimation with Insole-Based Sensor Fusion}

\author{
Liang Yu\,\orcidlink{0009-0001-6893-3961},
Tao Zhu\,\orcidlink{0000-0002-5879-5980},
\IEEEmembership{Senior Member, IEEE},
Xiangqing Lu,
Wenjian Luo,
\IEEEmembership{Senior Member, IEEE},
and Liming Chen\,\orcidlink{0000-0003-0200-7989},
\IEEEmembership{Senior Member, IEEE}
\thanks{Manuscript received XXX XX, 2026; revised XXX XX, 2026.
This work was supported in part by XXX under Grant XXX.}
\thanks{Liang Yu, Tao Zhu, and Xiangqing Lu are with the University of South China,
Hengyang, Hunan, China
(e-mail: lyu@stu.usc.edu.cn; tzhu@usc.edu.cn; xqlu@stu.usc.edu.cn).}
\thanks{Wenjian Luo is with Harbin Institute of Technology, Harbin, China
(e-mail: luowenjian@hit.edu.cn).}
\thanks{Liming Chen is with Dalian University of Technology, Dalian, China
(e-mail: LimingChen0922@dlut.edu.cn).}
}

\maketitle

\begin{abstract}
Real-time monocular human pose estimation is an important sensing component for human-centered intelligent systems. While recent camera-based methods have achieved impressive accuracy, fundamental challenges persist, particularly when visual observations are incomplete, such as during lower-body occlusion or short out-of-frame intervals. In these challenging but common scenarios, vision-only systems often struggle to reconstruct physically plausible lower-limb motion and reliable foot-ground contact. Due to the highly unobtrusive and wearable nature of smart insoles, this paper investigates whether simple plantar pressure and acceleration sensors can effectively resolve these visually induced limitations. To systematically investigate this, we first collect a synchronized multimodal dataset and then propose a part-decoupled temporal fusion framework. The proposed model fuses foot-sensing features into the lower-body branch to compensate for incomplete visual evidence, while a lightweight coordinator exchanges information between the upper- and lower-body temporal representations to maintain whole-body consistency. The framework is evaluated with three representative real-time visual backbones based on 2D keypoints, 3D landmarks, and SMPL-based human mesh recovery. We evaluate our framework on three representative real-time visual backbones. Quantitative experiments demonstrate that incorporating simple foot sensors consistently improves pose accuracy, motion smoothness, and gait-related measurements over vision-only baselines, with the most significant gains observed under occlusion and short visual-dropout conditions. Ultimately, our results demonstrate that unobtrusive foot sensing provides vital and complementary physical observability when camera observations are temporarily unavailable.
\end{abstract}

\begin{IEEEkeywords}
Monocular pose estimation, plantar pressure, foot-mounted acceleration, multimodal sensor fusion, human motion estimation.
\end{IEEEkeywords}

\section{Introduction}
\label{sec:introduction}

\IEEEPARstart{R}{eal-time} human motion estimation is a crucial sensing capability for human-centered intelligent systems, ranging from industrial safety monitoring and ergonomic assessment to rehabilitation and human--machine interaction. Monocular camera-based pose estimation has become highly attractive for these applications due to its low deployment cost and the convenience of requiring only an ordinary RGB camera. Driven by deep learning, recent methods have made remarkable progress, enabling highly accurate and real-time estimation of 2D keypoints, 3D landmarks, and SMPL-based body representations from monocular video~\cite{fang2023alphapose,lu2024rtmo,zhao2023poseformerv2,zhu2023motionbert,fang2024spgformer,goel2023humans,shin2024wham}.

Despite this impressive progress, real-time monocular pose estimation remains fundamentally constrained. The most critical challenges emerge when visual observations are incomplete, such as during lower-body occlusion (e.g., hidden behind a desk) or short out-of-frame intervals (visual dropouts). Under these conditions, the underlying depth ambiguity and scale uncertainty of monocular vision become severely amplified~\cite{gong2023diffpose,sun2024repose}. Consequently, camera-only systems often produce inaccurate foot trajectories, unreliable contact states, and physically implausible lower-limb motion. This fundamental limitation hinders their reliability in contact-sensitive applications like motion-analysis tasks that strictly depend on foot-ground interaction, stance phase, cadence, and load transfer~\cite{shin2024wham,zhang2024mmvp}.

To address this issue, we investigate an elegant and practical solution: incorporating simple plantar pressure and acceleration sensors via smart insoles. In particular, we focus on a four-point pressure-inertial sensing configuration (comprising four pressure points and a foot-mounted accelerometer) as an unobtrusive instantiation of foot sensing. Unlike heavy and highly restrictive motion capture suits, smart insoles are highly lightweight and non-intrusive. They seamlessly integrate into daily footwear while continuously providing direct physical measurements of foot-ground contact, load transfer, and lower-limb dynamics. These signals are highly complementary to monocular visual features and, crucially, remain fully informative when visual cues are occluded or temporarily unavailable~\cite{wu2024soleposer,gao2024pressinpose,ying2025form}. Therefore, rather than replacing the camera, our goal is to investigate whether the simple addition of unobtrusive foot sensors can provide the necessary physical constraints to overcome the accuracy and stability bottlenecks of monocular pose estimation under visually degraded conditions.

To systematically study this, we first collected a new synchronized Vision--Plantar-Pressure--Acceleration dataset that covers diverse daily activities, along with specifically designed occlusion and visual-dropout scenarios. Based on this dataset, we propose a part-decoupled temporal fusion framework for insole-enhanced monocular pose estimation. Recognizing that insole signals are physically tied to lower-body motion, our design mainly drives the upper-body branch using visual features, while explicitly fusing visual, pressure, and acceleration features in the lower-body branch. A lightweight global coordinator is introduced to exchange information between the upper and lower temporal memories, ensuring that the final reconstructed motion maintains whole-body kinematic consistency. This architecture strategically applies foot-sensing information where it is most needed without overriding visually dominant body parts~\cite{zhang2024dynaip,gao2024pressinpose}.

\begin{figure*}[t]
    \centering
    \includegraphics[width=\textwidth]{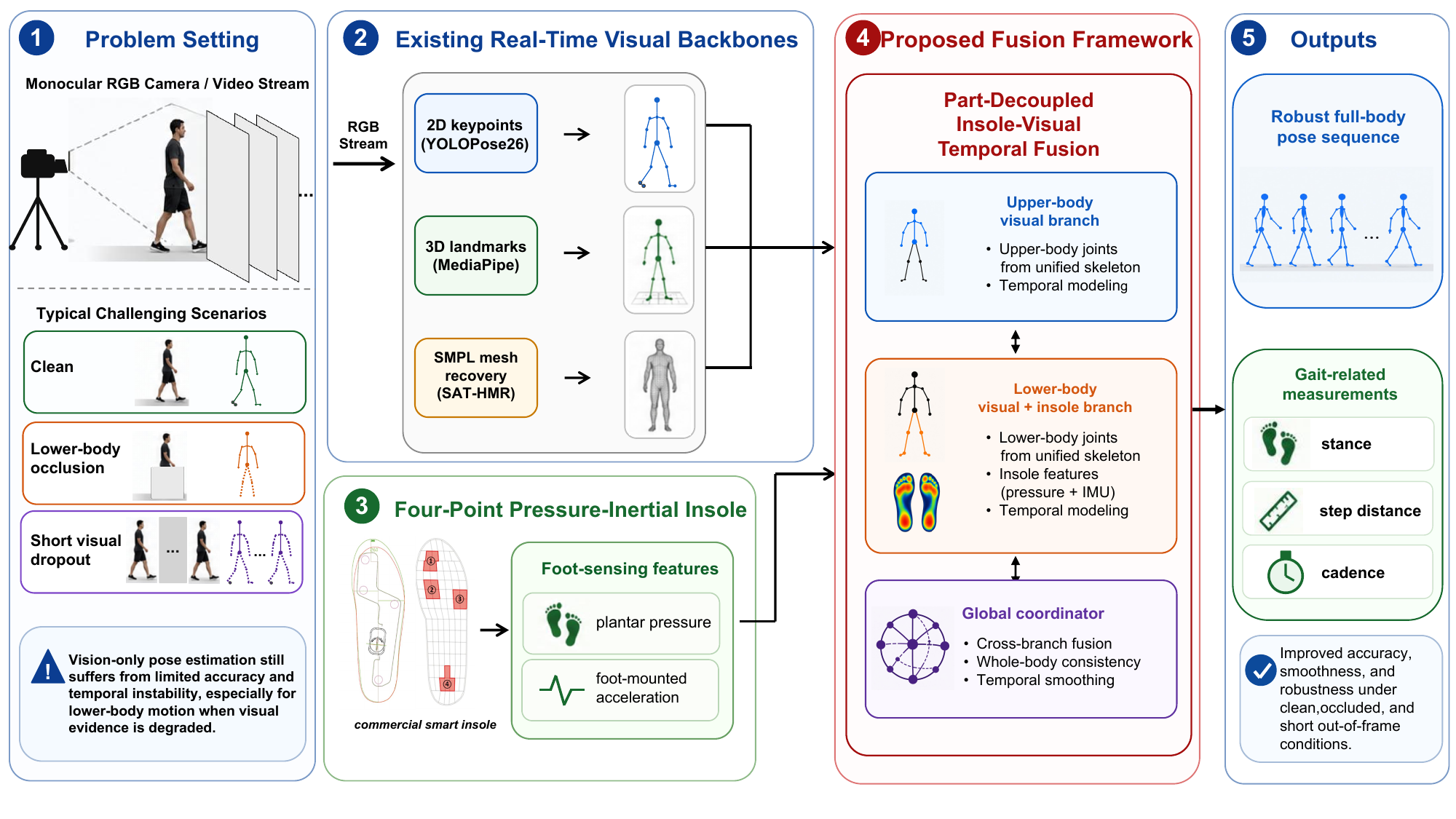}
    \caption{Overview of the proposed insole-enhanced real-time monocular human pose estimation framework. The system integrates three representative visual backbones (2D keypoints, 3D landmarks, and SMPL mesh recovery) with a four-point pressure-inertial smart insole via a part-decoupled temporal fusion module, producing robust full-body pose and gait-related outputs under clean, occluded, and short out-of-frame conditions.}
    \label{fig:overview}
\end{figure*}

We thoroughly evaluate the proposed method using three representative real-time monocular pose backbones. Experiments are conducted under clean visual conditions, lower-body occlusion, and short out-of-frame intervals. By evaluating standard pose errors, temporal smoothness, and gait-related measurements (stance accuracy, step-distance error, and cadence error), we prove that the reconstructed motion becomes highly reliable for contact-sensitive motion analysis~\cite{zhang2024mmvp,xu2026opensim}. The comprehensive results confirm a complete story: unobtrusive foot sensing effectively addresses the inherent blind spots of monocular cameras.

The main contributions of this paper are summarized as follows:
\begin{itemize}
    \item We experimentally demonstrate that plantar-pressure and foot-mounted acceleration signals can enhance real-time monocular human pose estimation under both normal and degraded visual conditions. The improvement is particularly evident under lower-body occlusion and short out-of-frame visual degradation, indicating that foot-sensing signals provide complementary physical cues beyond monocular visual observations.
        
    \item We propose a new vision--plantar-pressure--acceleration dataset for robust monocular pose estimation. The dataset contains synchronized monocular video, four-point plantar-pressure signals, foot-mounted acceleration, and Xsens reference motion, covering approximately 890K synchronized frames with five additional visual-degradation actions.
        
    \item We propose a part-decoupled temporal fusion framework that incorporates plantar-pressure and foot-acceleration cues into lower-body motion estimation while maintaining whole-body kinematic consistency through a lightweight global coordinator.
\end{itemize}

\section{Related Work}

\subsection{Real-Time Monocular Human Pose Estimation}

Monocular human pose estimation has been widely studied for low-cost and deployment-friendly motion sensing. Existing methods can be roughly grouped into 2D keypoint estimation, 3D landmark estimation, and human mesh recovery. For 2D pose estimation, YOLO-based, top-down, and whole-body pose estimators have improved real-time inference and keypoint localization accuracy~\cite{Maji_2022_CVPR,jiang2023rtmpose,lu2024rtmo,fang2023alphapose,yang2023dwpose,xu2022vitpose}. Lightweight landmark systems such as BlazePose further support on-device human pose tracking for mobile and interactive applications~\cite{bazarevsky2020blazepose}. For monocular 3D pose estimation, recent temporal models exploit transformers, frequency-domain representations, or graph-based reasoning to improve video-based 3D pose stability~\cite{zhao2023poseformerv2,zhu2023motionbert,mehraban2024motionagformer}. Human mesh recovery methods estimate SMPL-like body representations and have advanced from image-based recovery to video-based and promptable reconstruction~\cite{cai2024playing,goel2023humans,shin2024wham,wang2025prompthmr,Su_2025_CVPR}. However, these methods mainly rely on visual observations and learned pose priors. When lower-body evidence is occluded, noisy, or temporarily missing, camera-only systems may still suffer from unstable foot motion and physically implausible lower-limb reconstruction.

\begin{figure*}[t]
    \centering
    \includegraphics[width=\textwidth]{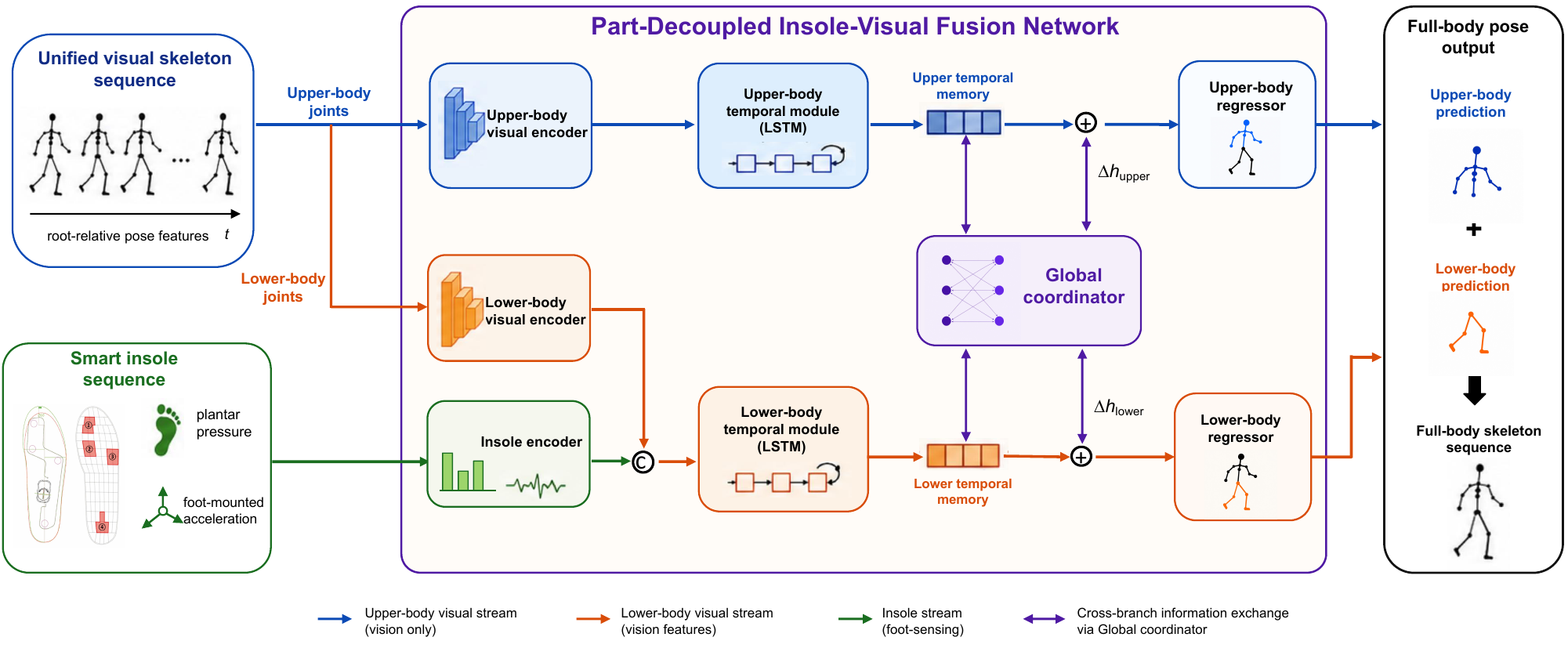}
    \caption{Part-decoupled vision--pressure--inertial fusion network. Visual pose features are separated into upper- and lower-body streams, while plantar-pressure and foot-acceleration features are fused into the lower-body stream. A global coordinator exchanges information between upper and lower temporal memories to maintain whole-body consistency.}
    \label{fig:network}
\end{figure*}

\subsection{Foot-Sensing-Based Human Motion Estimation}

Foot sensing provides complementary physical information for human motion estimation. Plantar pressure reflects contact state, center-of-pressure variation, load transfer, and gait phase, while foot-mounted inertial signals capture short-term lower-limb dynamics. Wearable pressure sensors, smart shoes, and IMU-based systems have therefore been widely explored for gait analysis, locomotion recognition, and biomechanical motion assessment~\cite{peng2024plantar,weigand2022imu,xu2026opensim,li2025p2lhap,li2025harmamba}. Recent studies further show that foot sensing can support full-body pose or motion reconstruction. SolePoser estimates full-body pose from a pair of insole sensors, PressInPose integrates pressure and inertial sensing for activity-level pose estimation, and FoRM reconstructs SMPL motion from dual-modal plantar pressure and inertial signals~\cite{wu2024soleposer,gao2024pressinpose,ying2025form}. Sparse-IMU and mobile-device-based systems also demonstrate the potential of compact wearable sensing for full-body motion estimation~\cite{zhang2024dynaip,xu2024mobileposer}. Nevertheless, foot-only or sparse-sensor-only reconstruction is inherently under-constrained because upper-body posture and arm motion cannot be directly observed from foot signals alone. Our work therefore studies plantar-pressure and foot-acceleration signals as complementary physical evidence for monocular pose estimation rather than as a replacement for vision.

\subsection{Vision and Foot-Sensing Fusion for Human Motion Capture}

A closely related direction is to incorporate foot-ground interaction into vision-based motion capture. Physics-aware methods use contact, center-of-pressure, or body-ground constraints to reduce foot sliding and improve physically plausible pose estimation~\cite{tripathi2023physics,shin2024wham}. Multimodal datasets and systems such as MMVP and MotionPRO further demonstrate that pressure information can improve visual motion capture, especially for foot contact, global translation, and human--world interaction modeling~\cite{zhang2024mmvp,ren2025motionpro}. These studies confirm the value of pressure and contact cues for vision-based human motion estimation. However, many existing works focus on dataset construction, offline fitting, pressure-mat settings, or dedicated multimodal MoCap pipelines. In contrast, this paper targets real-time monocular pose estimation and studies whether four-point plantar pressure and foot-mounted acceleration can improve real-time monocular pose estimation under incomplete visual observations.

\section{Vision--Plantar-Pressure--Acceleration Dataset}
\label{sec:dataset}

To support the study of insole-enhanced monocular pose estimation, we introduce a comprehensive and synchronized Vision--Plantar-Pressure--Acceleration dataset. The dataset contains synchronized monocular RGB video, four-point plantar-pressure signals, foot-mounted acceleration, and high-quality reference motion annotations. 

\begin{table*}[t]
\centering
\caption{Action composition of the new vision--plantar-pressure--acceleration dataset. Detailed action definitions and preprocessing scripts will be provided in the project repository.}
\label{tab:action_composition}
\footnotesize
\setlength{\tabcolsep}{3pt}
\renewcommand{\arraystretch}{1.18}
\begin{tabular*}{\textwidth}{@{\extracolsep{\fill}}p{0.18\textwidth} c c p{0.38\textwidth} p{0.18\textwidth}@{}}
\toprule
Subset & Actions & Frames & Representative actions & Usage \\
\midrule
Home activities 
& 13 
& 338K 
& standing, dressing, brushing teeth, washing face, eating, tying shoelaces, using phone, hanging laundry, watching TV, sweeping, mopping
& Standard evaluation \\

Office activities 
& 5 
& 92K 
& desk work, getting water, drinking water, restroom use, crossing legs
& Standard evaluation \\

Sports/exercise motions 
& 12 
& 212K 
& marching, walking, jogging, cycling, stairs, jumping rope, tennis, badminton, basketball, kicking, dumbbell exercises, dancing
& Standard evaluation \\

Rehabilitation-related motions 
& 8 
& 208K 
& alternating lunges, squats, horse stance, turning, jumping, sit-to-stand, jumping jacks, tiptoeing
& Standard evaluation \\
\midrule
Short out-of-frame actions 
& 3 
& 25K 
& jumping out and returning; walking out and returning; leaving the view, sitting, and returning
& Visual-dropout analysis \\

Desk-induced real occlusion 
& 2 
& 15K 
& upper-body free motion with static lower body; walking with lower body fully occluded by a desk
& Real-occlusion stress test \\
\midrule
Total 
& 43 
& 890K 
& normal, short out-of-frame, and desk-induced real-occlusion actions
& Full dataset \\
\bottomrule
\end{tabular*}
\end{table*}

\subsection{Data Acquisition and Synchronization}
The data collection platform consists of three synchronized subsystems:
\begin{enumerate}
    \item \textbf{Smart Insoles}: A pair of commercial I-SOL smart insoles, each of which is equipped with four piezoresistive pressure sensors and a 3-axis accelerometer and samples data at 30 Hz and transmits it via Bluetooth.
    \item \textbf{Monocular Camera}: An RGB camera is positioned frontally to ensure all full-body details are captured, recording monocular video at 120 Hz (downsampled to 30 Hz).
    \item \textbf{Reference Motion Capture}: An Xsens Awinda inertial motion capture system with 17 high-fidelity sensors captures ground-truth full-body kinematics at 60 Hz (downsampled to 30 Hz).
\end{enumerate}
The camera and the Xsens system are connected to the same workstation, enabling direct and precise software synchronization. The wireless insole stream is then aligned to this shared timeline by computing a correlation offset from the accelerometer spikes during a ``marching in place'' calibration routine at the beginning of each session.

\subsection{Action Composition and Protocol}
The dataset comprises approximately 890K synchronized frames from 20 healthy adult participants (6 female, 14 male; mean age: 24 years). \textbf{All experimental protocols involving human subjects were approved by the local ethics committee, and written informed consent was obtained from all participants prior to the experiments.} It covers 43 distinct activities categorized into six subsets, as detailed in Table~\ref{tab:action_composition}.

The recordings are divided into standard evaluation sets (Home, Office, Sports, and Rehabilitation) and visual-degradation sets. The visual-degradation actions include short out-of-frame dropouts (for qualitative availability analysis) and desk-induced real occlusions (for physical obstruction stress testing). To prevent long-horizon inertial drift from Xsens, the dataset collection follows a block-wise protocol, limiting each continuous recording segment to 10--15 minutes with repeated recalibration.

\section{Method}
\label{sec:method}

We propose an insole-enhanced framework for real-time monocular human pose estimation (Fig.~\ref{fig:overview}). The system takes the standardized output of a real-time visual backbone $\mathbf{v}_{1:T}^{b}$ and a synchronized smart-insole stream $\mathbf{s}_{1:T}$ as inputs to predict a 22-joint SMPL-style root-relative 3D skeleton $\hat{\mathbf{P}}_{1:T} = f_{\theta}(\mathbf{v}_{1:T}^{b}, \mathbf{s}_{1:T})$, where $\hat{\mathbf{P}}_{t} \in \mathbb{R}^{J \times 3}$ and $J=22$.

\subsection{Part-Decoupled Vision--Pressure--Inertial Fusion Network}

The proposed fusion network is motivated by the physical relationship between the insole signal and the human body. Since pressure and acceleration from the feet mainly describe lower-body contact and dynamics, directly concatenating them with the whole-body visual representation may be suboptimal. We therefore use a part-decoupled design, as illustrated in Fig.~\ref{fig:network}.

The visual feature is encoded into upper-body and lower-body streams via linear projection networks (MLPs) denoted as $E_{u}$ and $E_{l}^{v}$:
\begin{equation}
    \mathbf{h}_{t}^{u} = E_{u}(\mathbf{v}_{t}^{b}), \quad \mathbf{h}_{t}^{l,v} = E_{l}^{v}(\mathbf{v}_{t}^{b}).
\end{equation}
The insole feature is encoded separately:
\begin{equation}
    \mathbf{h}_{t}^{s} = E_{s}(\mathbf{s}_{t}).
\end{equation}
The lower-body stream then fuses visual and insole features:
\begin{equation}
    \mathbf{h}_{t}^{l} = E_{l}^{f}([\mathbf{h}_{t}^{l,v}; \mathbf{h}_{t}^{s}]),
\end{equation}
where $[\cdot;\cdot]$ denotes concatenation. This design injects foot-sensing cues into the body part where they are physically most informative, while keeping the upper-body stream mainly vision-driven.

The upper and lower streams are modeled by separate temporal modules:
\begin{equation}
    \mathbf{m}_{1:T}^{u} = \mathrm{LSTM}_{u}(\mathbf{h}_{1:T}^{u}), \quad \mathbf{m}_{1:T}^{l} = \mathrm{LSTM}_{l}(\mathbf{h}_{1:T}^{l}).
\end{equation}

\begin{table*}[t]
\centering
\caption{Main pose estimation results under clean, lower-body occlusion, and visual dropout conditions. MPJPE, PA-MPJPE, and jitter are lower-is-better.}
\label{tab:main_results}
\footnotesize
\setlength{\tabcolsep}{5pt}
\renewcommand{\arraystretch}{1.25} 
\begin{tabular*}{\textwidth}{@{\extracolsep{\fill}}llllccc@{}}
\toprule 
Condition & Evaluation Scope & Backbone & Method
& MPJPE (mm) $\downarrow$
& PA-MPJPE (mm) $\downarrow$
& Jitter ($\mathrm{m/s^3}$) $\downarrow$ \\
\midrule 
\multirow{6}{*}{Clean}
& \multirow{6}{*}{Standard Subsets} & MediaPipe & Vision Temporal & 67.55 & 45.21 & 81.40 \\
& & MediaPipe & Ours & \textbf{64.15} & \textbf{40.62} & \textbf{65.05} \\
& & YOLOPose26 & Vision Temporal & 68.13 & 45.37 & 80.28 \\
& & YOLOPose26 & Ours & \textbf{63.94} & \textbf{41.57} & \textbf{69.35} \\
& & SAT-HMR & Vision Temporal & 63.96 & 42.64 & 68.29 \\
& & SAT-HMR & Ours & \textbf{58.38} & \textbf{38.06} & \textbf{66.11} \\
\midrule 
\multirow{6}{*}{Occlusion}
& \multirow{6}{*}{Standard Subsets} & MediaPipe & Vision Temporal & 102.89 & 65.73 & 183.12 \\
& & MediaPipe & Ours & \textbf{68.59} & \textbf{43.72} & \textbf{82.65} \\
& & YOLOPose26 & Vision Temporal & 112.38 & 76.70 & 110.30 \\
& & YOLOPose26 & Ours & \textbf{66.35} & \textbf{42.78} & \textbf{79.03} \\
& & SAT-HMR & Vision Temporal & 76.75 & 50.88 & 93.38 \\
& & SAT-HMR & Ours & \textbf{61.84} & \textbf{41.28} & \textbf{81.95} \\
\midrule 
Dropout & Dropout Subset & --- & Ours & \textbf{123.63} & \textbf{59.65} & \textbf{105.15} \\
\bottomrule 
\end{tabular*}
\end{table*}

While the decoupled streams prevent local foot-sensing noise from overriding upper-body visual observations, completely isolating the two branches violates the kinetic chain principle of the human body. The pelvis and lower spine act as a mechanical bridge; lower-limb momentum generated during walking inherently influences torso rotation and arm swinging. To maintain this whole-body kinematic consistency, a lightweight global coordinator acts as a soft kinetic constraint. It cross-examines the hidden states of both temporal memories and generates residual updates:
\begin{equation}
    [\Delta \mathbf{m}_{t}^{u}; \Delta \mathbf{m}_{t}^{l}] = C_{\mathrm{MLP}}([\mathbf{m}_{t}^{u}; \mathbf{m}_{t}^{l}]).
\end{equation}
The coordinated memories are then used to regress upper-body and lower-body joints via MLP-based regressors $H, H_u$, and $H_l$:
\begin{equation}
    \hat{\mathbf{P}}_{t} = H\left(H_{u}(\mathbf{m}_{t}^{u}+\Delta\mathbf{m}_{t}^{u}), H_{l}(\mathbf{m}_{t}^{l}+\Delta\mathbf{m}_{t}^{l})\right).
\end{equation}
This structure allows the model to exploit insole cues for lower-body refinement while preserving full-body kinematic coherence.

\subsection{Training Objective}

The training objective is designed to supervise both frame-level pose accuracy and temporal motion quality. Since the output is a continuous root-relative skeleton sequence, only minimizing the joint position error may still lead to frame-wise fluctuations or locally distorted limb configurations. Following standard practices in temporal 3D human pose estimation~\cite{pavllo20193d,kocabas2020vibe,zhou2017towards}, we combine a joint position loss with first- and second-order temporal smoothness terms, and a bone-length consistency constraint:

\begin{equation}
    \mathcal{L} = \mathcal{L}_{pos} + \lambda_{v}\mathcal{L}_{vel} + \lambda_{a}\mathcal{L}_{acc} + \lambda_{b}\mathcal{L}_{bone}.
\end{equation}
Here, $\mathcal{L}_{pos}$ penalizes joint position errors in the root-relative skeleton space and provides the main supervision for pose accuracy. $\mathcal{L}_{vel}$ and $\mathcal{L}_{acc}$ constrain first- and second-order temporal differences, respectively, encouraging the predicted motion to follow the reference dynamics and reducing high-frequency jitter. $\mathcal{L}_{bone}$ preserves local kinematic structure by penalizing deviations in bone lengths, which helps avoid implausible limb deformation during temporal fusion. In our implementation, $\lambda_{v}=0.1$, $\lambda_{a}=0.4$, and $\lambda_{b}=0.02$.

To improve robustness to temporary visual failures, we apply visual feature dropout during training. For selected training windows, a continuous temporal block of visual features is masked to zero while the synchronized insole stream remains available. The masked block length is randomly sampled within the sequence, so the model learns to use pressure and acceleration cues when visual observations are temporarily unreliable. This augmentation is used only during training.

\subsection{Real-Time Streaming Inference}

During inference, the proposed fusion module operates in a streaming manner. At each time step, it receives the current visual pose feature and the synchronized insole measurement, while maintaining recurrent hidden states from previous frames. The current pose is predicted as:
\begin{equation}
    \hat{\mathbf{P}}_{t} = f_{\theta}^{online}(\mathbf{v}_{t}^{b}, \mathbf{s}_{t}, \mathbf{h}_{t-1}),
\end{equation}
where $\mathbf{h}_{t-1}$ denotes the temporal memory from the previous frame. This design avoids reprocessing the full sequence and keeps the additional overhead of the insole-enhanced module small, making it compatible with real-time visual pose backbones.

\section{Experiments}
\label{sec:experiments}

\begin{table*}[t]
\centering
\caption{Gait-related motion parameters under clean, lower-body occlusion, and visual dropout conditions. Stance accuracy is higher-is-better, while step-distance error and cadence error are lower-is-better.}
\label{tab:gait_results}
\setlength{\tabcolsep}{4pt}
\renewcommand{\arraystretch}{1.25} 
\resizebox{\textwidth}{!}{
\begin{tabular}{l l l l ccc}
\toprule 
Condition & Evaluation Scope & Backbone & Method
& Stance Acc. (\%) $\uparrow$
& Step Dist. Err. (cm) $\downarrow$
& Cadence Err. (step/min) $\downarrow$ \\
\midrule 
\multirow{6}{*}{Clean}
& \multirow{6}{*}{Walking \& Jogging} & MediaPipe & Vision Temporal & 74.15 & 8.34 & 11.42 \\
& & MediaPipe & Ours & \textbf{79.78} & \textbf{5.54} & \textbf{6.23} \\
& & YOLOPose26 & Vision Temporal & 67.89 & 6.69 & 16.21 \\
& & YOLOPose26 & Ours & \textbf{78.16} & \textbf{4.74} & \textbf{4.15} \\
& & SAT-HMR & Vision Temporal & 66.95 & 6.91 & 10.50 \\
& & SAT-HMR & Ours & \textbf{78.03} & \textbf{5.56} & \textbf{8.36} \\
\midrule 
\multirow{6}{*}{Occlusion}
& \multirow{6}{*}{Walking \& Jogging} & MediaPipe & Vision Temporal & 62.80 & 14.92 & 15.17 \\
& & MediaPipe & Ours & \textbf{80.15} & \textbf{5.73} & \textbf{6.23} \\
& & YOLOPose26 & Vision Temporal & 60.43 & 12.38 & 21.80 \\
& & YOLOPose26 & Ours & \textbf{77.69} & \textbf{5.65} & \textbf{3.11} \\
& & SAT-HMR & Vision Temporal & 62.86 & 11.84 & 24.57 \\
& & SAT-HMR & Ours & \textbf{76.99} & \textbf{6.93} & \textbf{6.42} \\
\midrule
Dropout & Dropout Walking & --- & Ours & \textbf{74.08} & \textbf{9.55} & \textbf{12.37} \\
\bottomrule
\end{tabular}
}
\end{table*}

\subsection{Experimental Setup}
We evaluate our method on the new VPPA dataset (Section \ref{sec:dataset}). Three representative real-time visual backbones are evaluated: MediaPipe, YOLOPose26, and SAT-HMR. Reference body poses are obtained from Xsens MVN. All compared methods use a fixed action-disjoint split where test actions are unseen during training. For synthetic lower-body occlusion, we manually mask the bottom 30\% of each video frame with black pixels before feature extraction. Training and validation use sliding windows of $T=120$ frames with a step size of 5 frames.

\subsection{Baselines and Compared Variants}

We compare the proposed method with a vision-only temporal baseline and several fusion variants. The vision-only temporal baseline uses the same visual backbone output and temporal modeling strategy as our method, but removes the insole stream. This baseline is important because it separates the benefit of pressure--inertial sensing from the benefit of temporal smoothing alone. For each visual backbone, the baseline is trained and evaluated in the same root-relative 22-joint output space as the proposed model.

We also evaluate a naive temporal fusion variant, which directly concatenates visual and insole features before temporal modeling without part decoupling or cross-branch coordination. This variant tests whether simple multimodal concatenation is sufficient. In addition, we report ablations without the global coordinator, without pressure, and without acceleration. These variants examine the contribution of cross-branch information exchange and the two sensing modalities. All compared methods use the same data split, window length, training objective, and evaluation metrics for a fair comparison.

\subsection{Implementation Details}

All models are trained on 120-frame windows. The training and validation windows are sampled with a step size of 5 frames, while testing uses a step size of 1 frame to obtain dense frame-level predictions. The proposed fusion model uses two-layer LSTM temporal modules with a hidden dimension of 256. The upper- and lower-body branches use separate temporal memories, and the global coordinator predicts residual updates for both branches. We train the model using AdamW with cosine learning-rate scheduling and select the best checkpoint according to validation PA-MPJPE. For fair comparison, the vision-only temporal baseline and fusion variants are trained under the same data split, window length, and evaluation protocol.

\subsection{Evaluation Metrics}

We report MPJPE and PA-MPJPE for pose accuracy, and jerk-based jitter for temporal smoothness. MPJPE measures the mean Euclidean distance between the predicted and reference joints in the root-relative skeleton space, while PA-MPJPE is computed after Procrustes alignment. Jitter is computed from the third-order temporal derivative of the predicted joint trajectories and reflects high-frequency motion instability.

For gait-related evaluation, we report stance accuracy, step-distance error, and cadence error. These metrics are computed from the reconstructed foot trajectories and are used to examine whether the estimated pose preserves contact-sensitive motion properties beyond joint-level accuracy. 

\textbf{To ensure biomechanical validity and clinical relevance, gait-related metrics are evaluated exclusively on standard cyclic locomotion sequences within the test split (specifically, walking and jogging). This selective evaluation prevents the introduction of ill-defined noise from static or non-locomotive activities (such as sitting, eating, or upper-body exercises), where the concepts of stride geometry and gait cycle are physically inapplicable.}

Stance accuracy evaluates frame-level left/right foot-contact correctness. For each frame, we infer the predicted contact state from the reconstructed foot trajectory and compare it with the reference contact label derived from the motion-capture system. The smart-insole pressure signal is not used as the ground-truth contact label for this metric, preventing the evaluation from directly reusing the input sensing modality. 

Step-distance error measures the accuracy of step geometry. We first identify heel-strike events from the reference contact sequence. At each detected heel-strike frame, we compute the horizontal distance between the left and right ankle joints. The reported error is the mean absolute difference between the predicted and reference step distances. 

Cadence error measures the accuracy of temporal step rhythm. Cadence is computed as the number of detected steps per minute over an evaluated locomotion sequence. Together, these gait-related metrics strictly evaluate whether the reconstructed motion is useful for downstream gait and contact analysis.

\subsection{Main Pose Estimation Results}

Table~\ref{tab:main_results} reports the main pose estimation results across three evaluation scopes: clean visual conditions, lower-body occlusion, and visual dropout (out-of-frame). 

Under clean visual observations on standard subsets, the proposed method consistently improves MPJPE, PA-MPJPE, and jitter across MediaPipe, YOLOPose26, and SAT-HMR. This indicates that the insole stream provides useful complementary cues even when the visual input is not deliberately degraded. The improvement becomes more pronounced under lower-body occlusion. Since the bottom 30\% of the video frame is masked, the visual backbones receive weaker lower-body evidence, while the insole stream still provides pressure and acceleration cues related to foot-ground interaction. As a result, the proposed method substantially reduces pose error and temporal jitter compared with the vision-only temporal baselines.

Crucially, we also report quantitative results under the extreme visual-dropout condition (Table~\ref{tab:main_results}, bottom row). During visual dropouts, the subject temporarily leaves the camera view. Because vision-only baselines completely lose physical observability and degenerate to static holding or unpredictable extrapolation without image evidence, they cannot serve as a meaningful comparative baseline (denoted as `---`). In stark contrast, the proposed insole-enhanced method reliably leverages continuous plantar pressure and acceleration to maintain active tracking of the lower limbs, achieving an MPJPE of 123.63 mm and a PA-MPJPE of 59.65 mm. This further supports our core claim: unobtrusive foot sensing provides critical physical constraints exactly when monocular cameras fail.

\begin{figure}[t] 
    \centering
    \setlength{\tabcolsep}{1pt} 
    \begin{tabular}{cccc}
        \makebox[0.24\columnwidth][c]{\footnotesize\textbf{Marching}} & 
        \makebox[0.24\columnwidth][c]{\footnotesize\textbf{Jogging}} & 
        \makebox[0.24\columnwidth][c]{\footnotesize\textbf{Jumping jacks}} & 
        \makebox[0.24\columnwidth][c]{\footnotesize\textbf{Alternating lunges}} \\[1.5mm]
        
        \includegraphics[width=0.17\columnwidth]{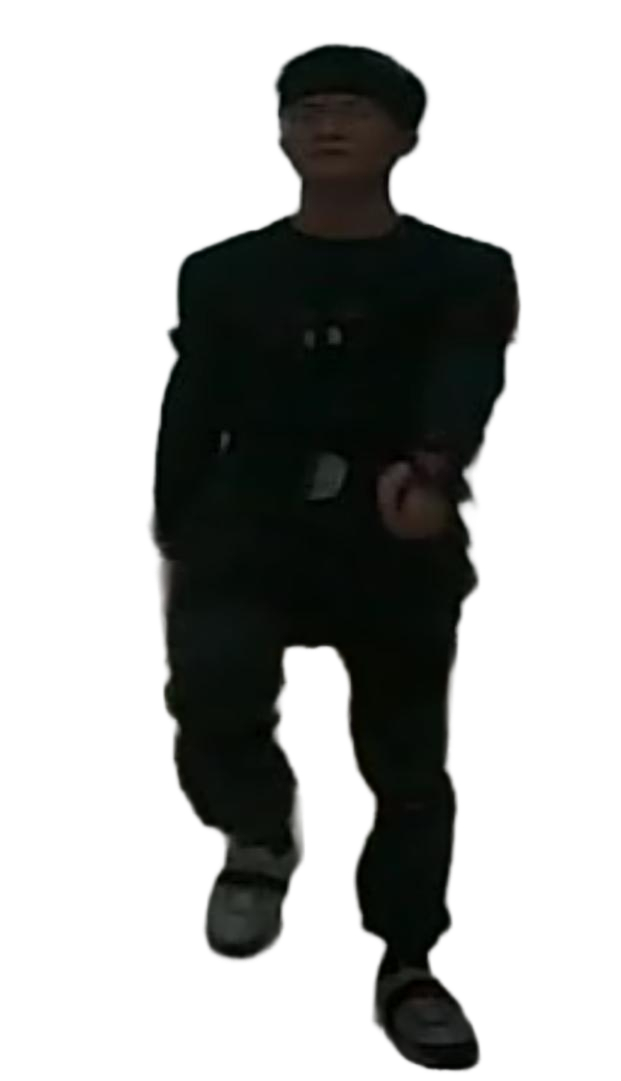} &
        \includegraphics[width=0.17\columnwidth]{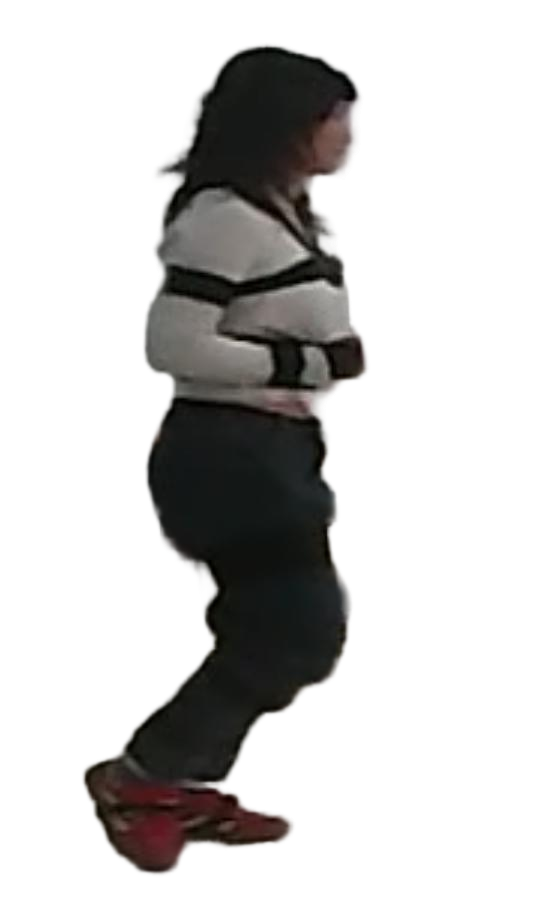} &
        \includegraphics[width=0.17\columnwidth]{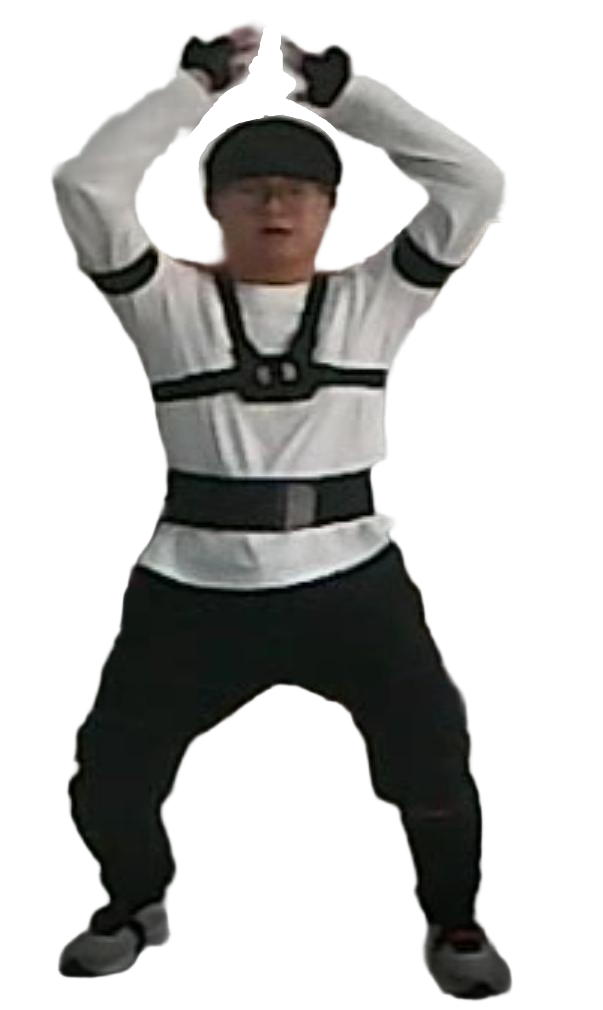} &
        \includegraphics[width=0.28\columnwidth]{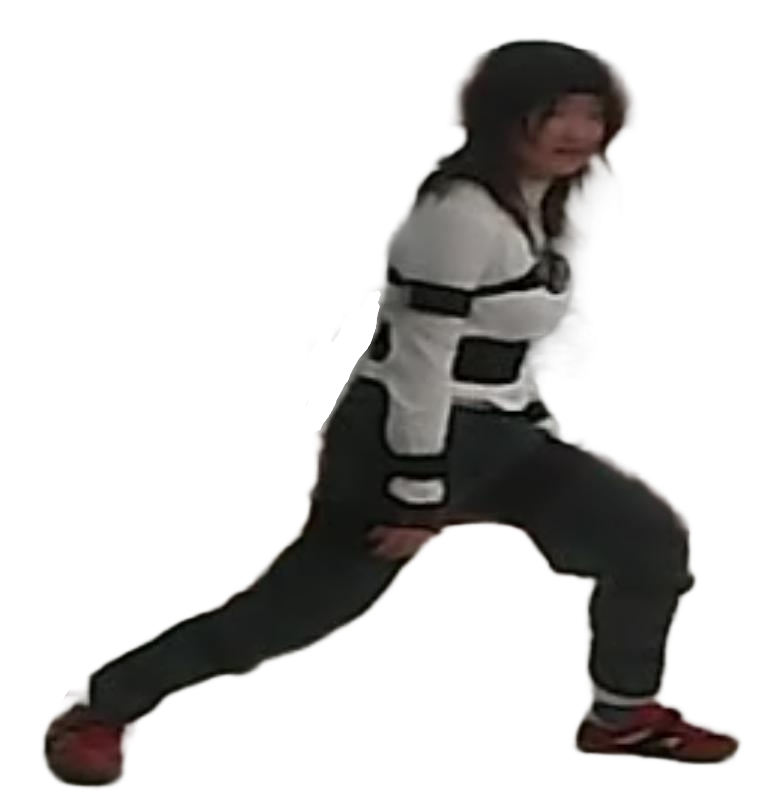} \\[-4mm] 
        
        \makebox[0.24\columnwidth][c]{\includegraphics[width=0.32\columnwidth]{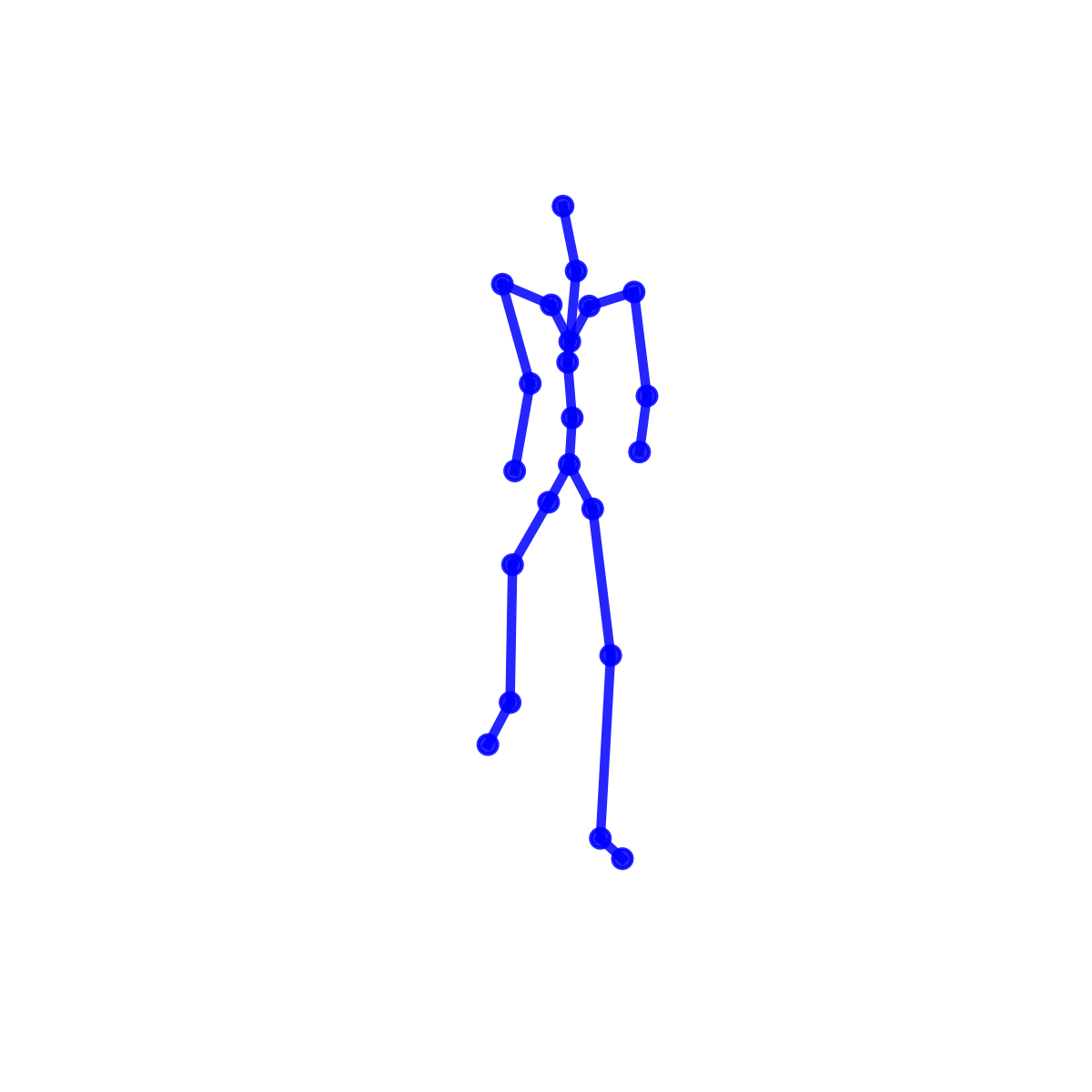}} &
        \makebox[0.24\columnwidth][c]{\includegraphics[width=0.32\columnwidth]{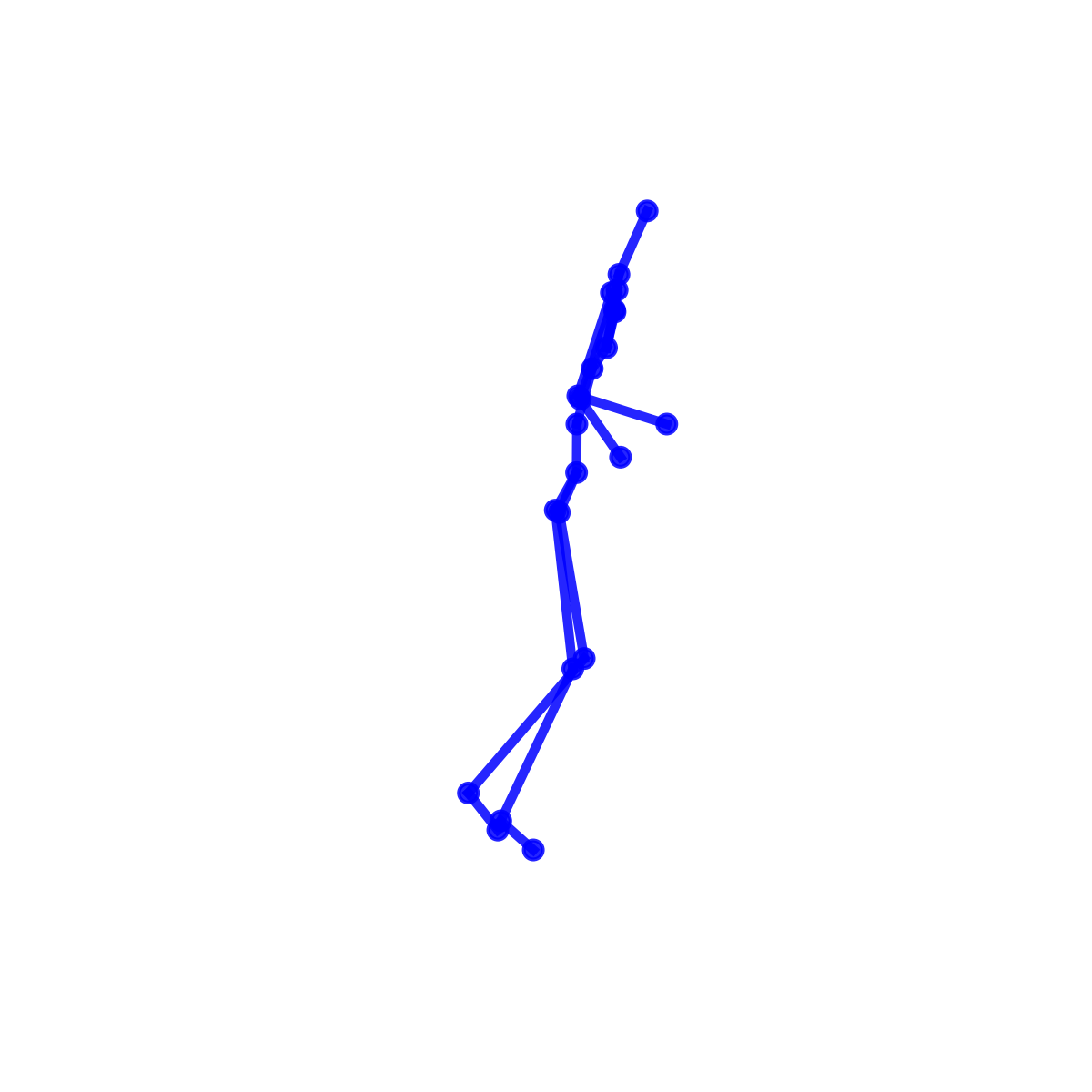}} &
        \makebox[0.24\columnwidth][c]{\includegraphics[width=0.32\columnwidth]{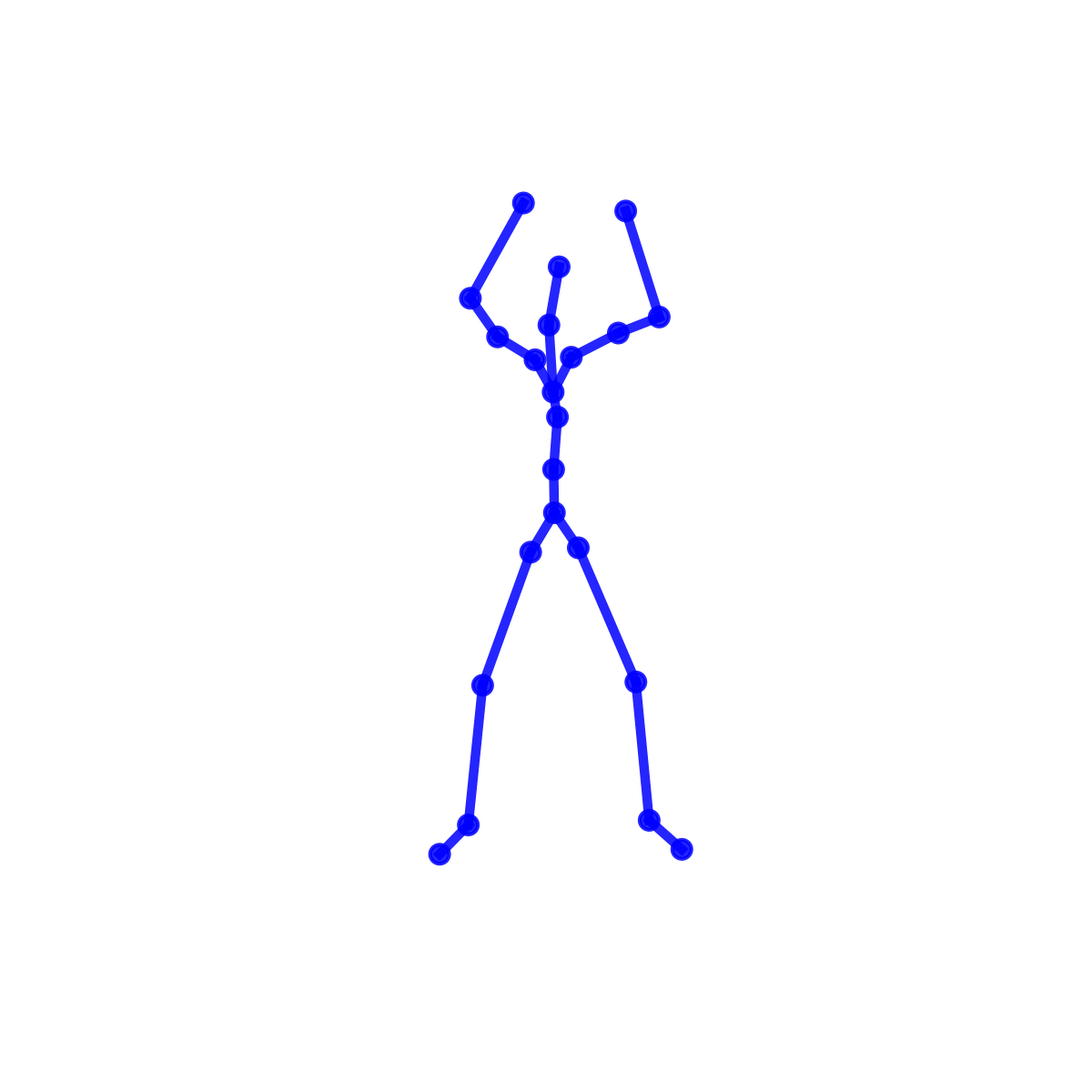}} &
        \makebox[0.24\columnwidth][c]{\includegraphics[width=0.32\columnwidth]{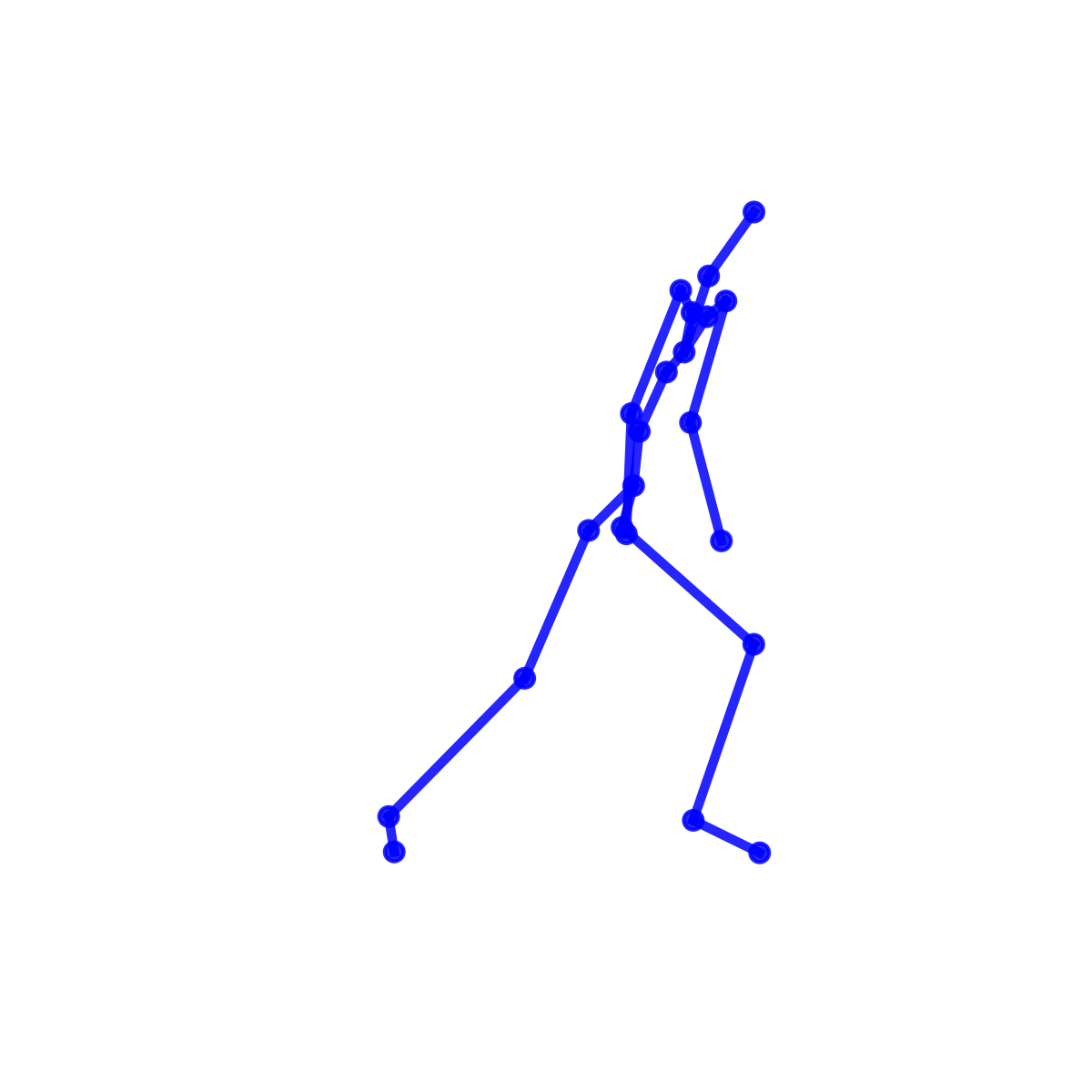}} \\[-8mm] 
        
        \makebox[0.24\columnwidth][c]{\includegraphics[width=0.32\columnwidth]{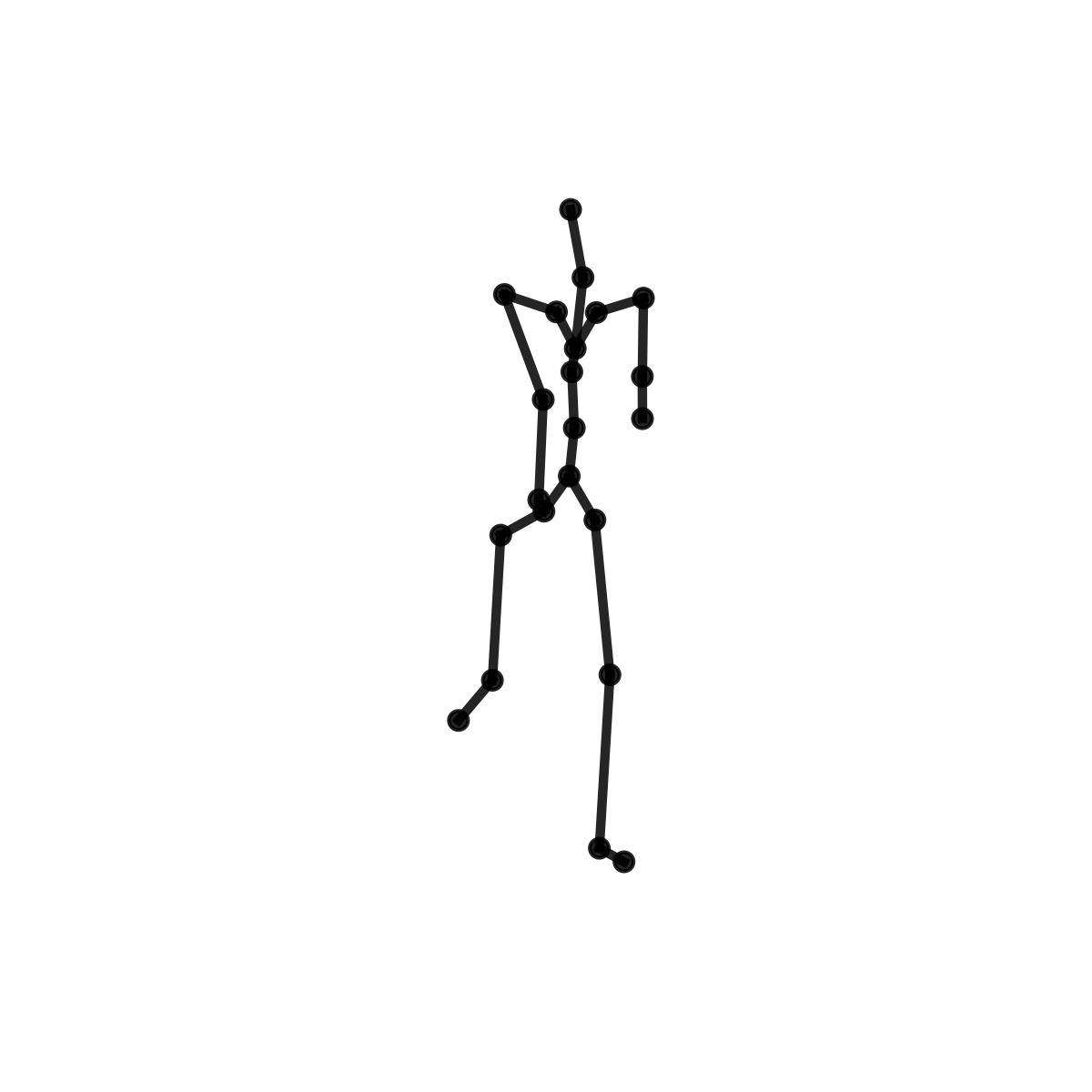}} &
        \makebox[0.24\columnwidth][c]{\includegraphics[width=0.32\columnwidth]{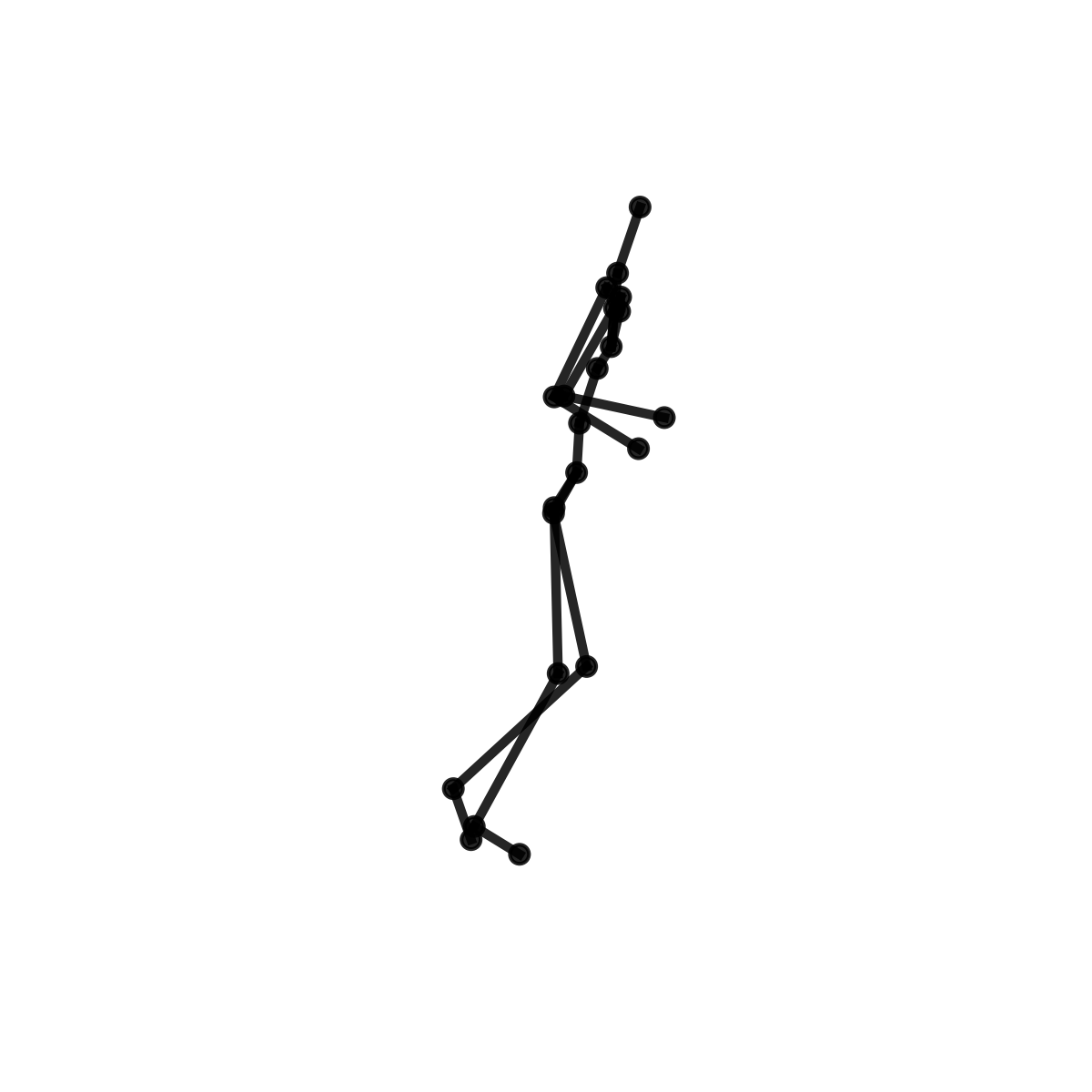}} &
        \makebox[0.24\columnwidth][c]{\includegraphics[width=0.32\columnwidth]{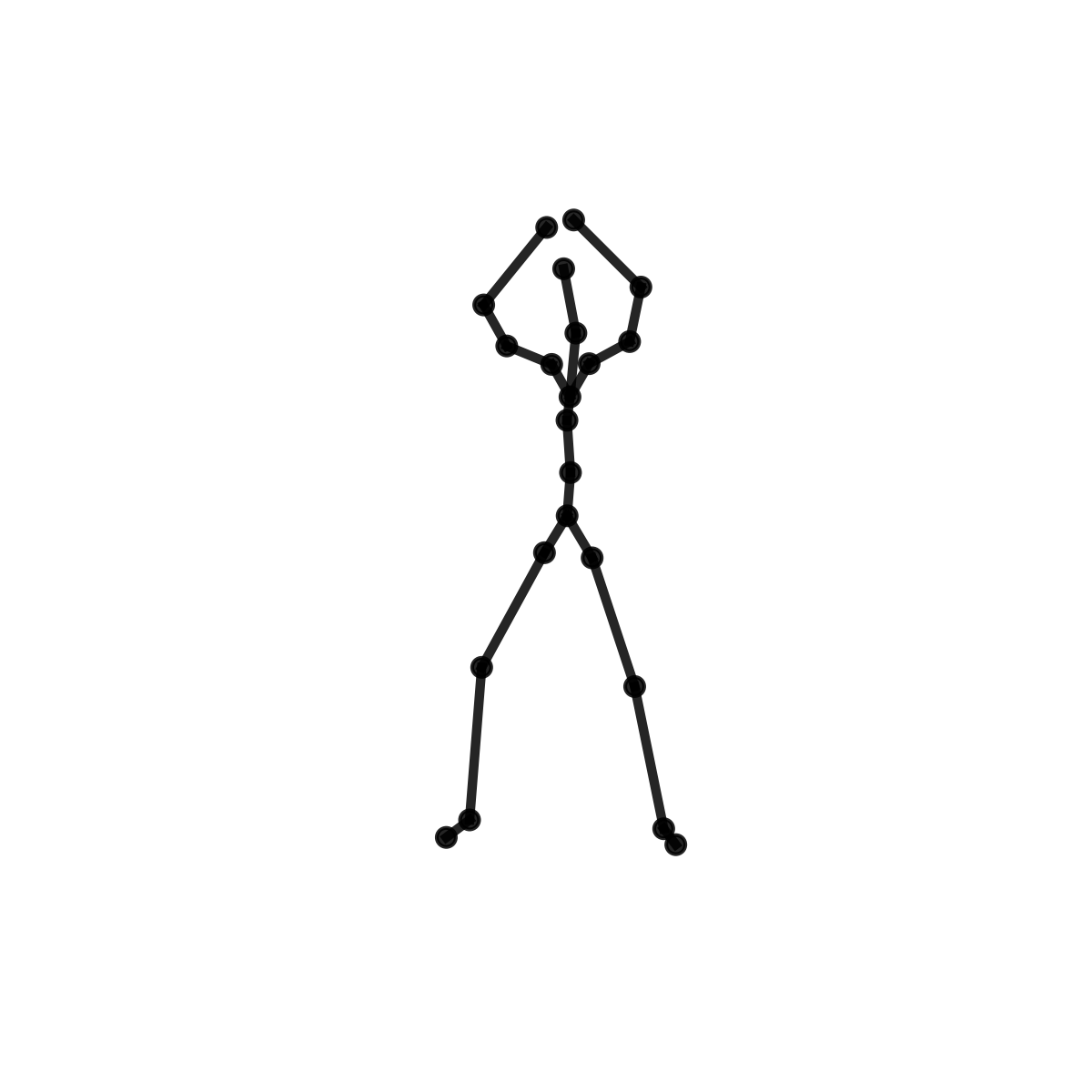}} &
        \makebox[0.24\columnwidth][c]{\includegraphics[width=0.32\columnwidth]{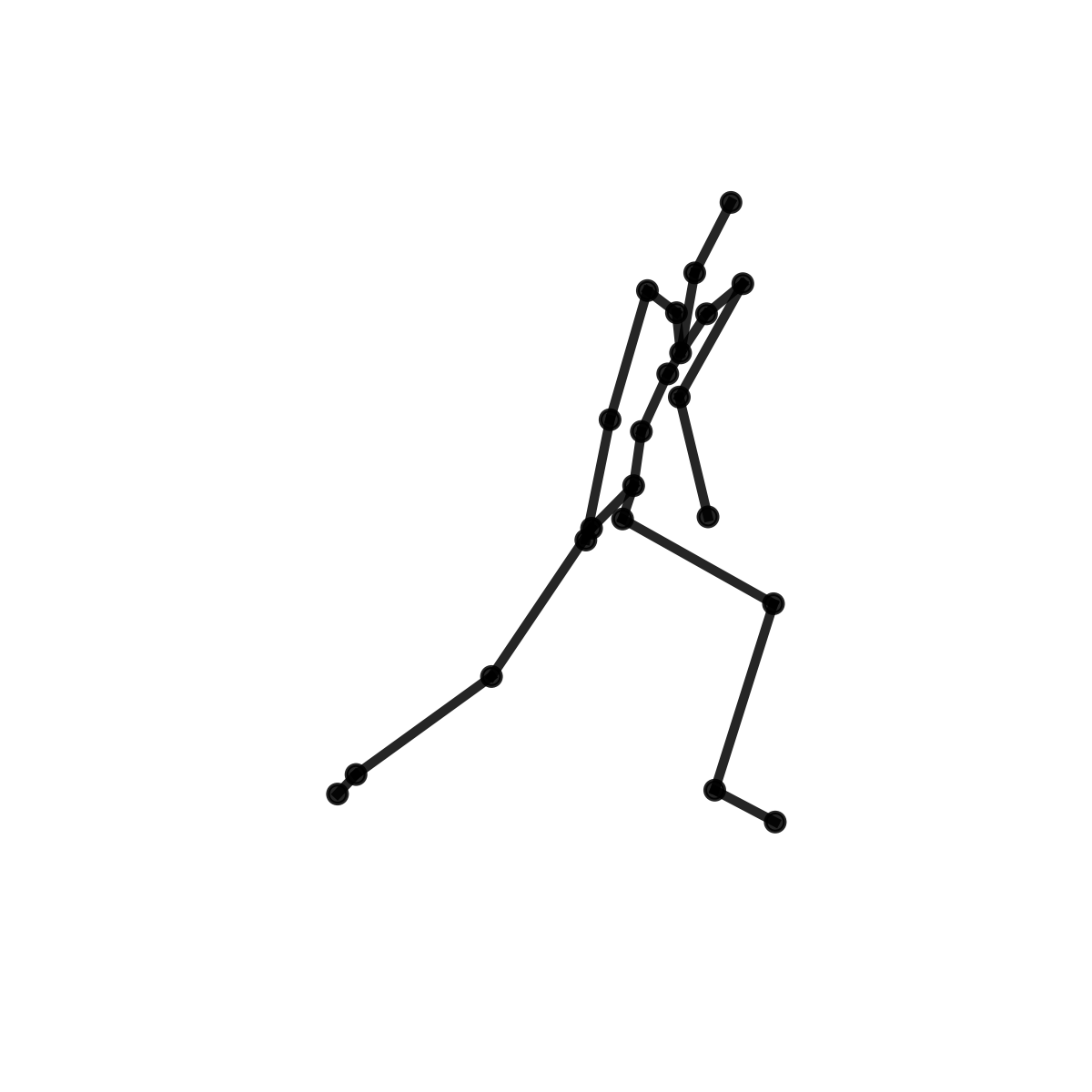}} \\
    \end{tabular}
    
    \vspace{-6mm} 
    \includegraphics[width=1\columnwidth]{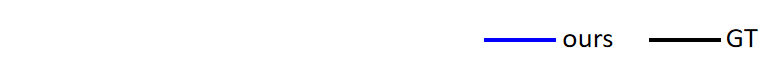} 
    \vspace{-5mm} 
    \caption{Qualitative comparison of full-body pose estimation on diverse highly-dynamic actions. Rows from top to bottom correspond to the input monocular images, poses estimated by our proposed method (Ours), and the Xsens reference ground truth, respectively.}
    \label{fig:qualitative_actions}
\end{figure}

\subsection{Gait-Related Evaluation}

Table~\ref{tab:gait_results} reports gait-related measurements specifically evaluated on cyclic locomotion (walking and jogging). 

In the clean setting, the proposed method improves stance accuracy for all three backbones and reduces both step-distance and cadence errors. These results indicate that the insole-enhanced model does not merely reduce average joint error, but fundamentally improves foot-contact timing, relative step geometry, and step rhythm. On average, stance accuracy increases by 8.99 percentage points, step-distance error decreases by 27.8\%, and cadence error decreases by 50.9\%.

The improvement is larger under lower-body occlusion. Vision-only baselines produce less reliable foot-contact and step-event estimates when lower-body observations are degraded. In contrast, the insole-enhanced model remains remarkably stable because plantar pressure and foot acceleration directly encode contact and foot dynamics. Under occlusion, the average stance accuracy increases by 16.25 percentage points. 

Furthermore, when evaluating locomotion during complete visual dropouts (Dropout Walking), the proposed method impressively maintains a stance accuracy of 74.08\%, a step-distance error of 9.55 cm, and a cadence error of 12.37 step/min. This demonstrates that even when the camera captures nothing, the smart insole continuously drives the temporal fusion network to output biomechanically plausible gait parameters.

\subsection{Cross-Subject Generalization}

To further examine whether the benefit of pressure--inertial foot sensing generalizes to unseen users, we conduct an additional subject-disjoint experiment using MediaPipe as the visual backbone. In this setting, models are trained on 15 subjects, while two subjects are used for validation and three held-out subjects are reserved purely for testing. We evaluate both clean and desk-occlusion conditions on the standard actions.

\begin{table}[t]
\centering
\caption{Cross-subject generalization results on held-out test subjects using MediaPipe. All metrics are lower-is-better.}
\label{tab:cross_subject}
\renewcommand{\arraystretch}{1.35} 
\setlength{\tabcolsep}{3pt} 
\resizebox{\columnwidth}{!}{
\begin{tabular}{llccc}
\toprule
Condition & Method & MPJPE (mm) $\downarrow$ & PA-MPJPE (mm) $\downarrow$ & Jitter ($\mathrm{m/s^3}$) $\downarrow$ \\
\midrule
\multirow{2}{*}{Clean}
& Temporal & 73.47 & 48.86 & 89.02 \\
& Ours & \textbf{68.06} & \textbf{45.78} & \textbf{75.64} \\
\midrule
\multirow{2}{*}{Occlusion}
& Temporal & 88.31 & 59.20 & 115.69 \\
& Ours & \textbf{74.48} & \textbf{50.43} & \textbf{83.21} \\
\bottomrule
\end{tabular}
}
\end{table}

As shown in Table~\ref{tab:cross_subject}, the proposed method still outperforms the vision-only temporal baseline on unseen subjects. Under clean conditions, it reduces MPJPE from 73.47 mm to 68.06 mm and decreases jitter from 89.02 to 75.64 $\mathrm{m/s^3}$. The advantage becomes larger under occlusion, where MPJPE is reduced from 88.31 mm to 74.48 mm and jitter is reduced from 115.69 to 83.21 $\mathrm{m/s^3}$. These results indicate that the benefit of pressure--inertial foot sensing is not limited to the calibrated-user action-disjoint setting, but can also improve robustness for held-out users.

\subsection{Real-Occlusion Stress Test}

We further evaluate the proposed method on two held-out real-occlusion sequences recorded with a physical desk. In both sequences, the desk is placed between the camera and the subject so that the lower body is fully invisible to the monocular camera. The first sequence is a stationary lower-body occlusion case, where the subject keeps the lower body nearly static while performing free upper-body movements. This setting tests whether the model can avoid visually induced lower-body drift when only upper-body motion is observable. The second sequence is a desk-occluded walking case, where the subject walks while the legs and feet remain hidden behind the desk. This setting is more challenging because the visual stream provides little direct evidence about step timing, foot displacement, or lower-limb articulation.

The synthetic and real occlusion settings serve different purposes. Synthetic lower-body occlusion masks the same actions used in the standard protocol, enabling controlled comparison with paired clean sequences. In contrast, desk-induced occlusion introduces physical scene-level obstruction, viewpoint-dependent visual ambiguity, and unpaired action sequences. Therefore, the real-occlusion actions are not used as a standard benchmark with clean counterparts, but as a stress test for practical obstacle-induced visual degradation.

\begin{table}[t]
\centering
\caption{Real-occlusion stress test on two held-out desk-induced occlusion sequences using MediaPipe. All metrics are lower-is-better.}
\label{tab:real_occlusion}
\renewcommand{\arraystretch}{1.35} 
\setlength{\tabcolsep}{3pt} 
\resizebox{\columnwidth}{!}{
\begin{tabular}{lccc}
\toprule
Method & MPJPE (mm) $\downarrow$ & PA-MPJPE (mm) $\downarrow$ & Jitter ($\mathrm{m/s^3}$) $\downarrow$ \\
\midrule
MediaPipe Temporal & 93.17 & 65.58 & 178.25 \\
Ours & \textbf{62.06} & \textbf{45.79} & \textbf{94.00} \\
\bottomrule
\end{tabular}
}
\end{table}

As shown in Table~\ref{tab:real_occlusion}, the proposed method remains substantially more robust than the vision-only temporal baseline under physical occlusion. It reduces MPJPE from 93.17 mm to 62.06 mm, PA-MPJPE from 65.58 mm to 45.79 mm, and jitter from 178.25 to 94.00 $\mathrm{m/s^3}$. These results complement the synthetic occlusion experiment: the synthetic setting provides controlled comparison under the same actions, while the real-occlusion setting verifies robustness under practical obstacle-induced visual degradation.

\begin{figure}[t]
\centering

\begin{tabular}{m{0.16\columnwidth} m{0.80\columnwidth}}
  
  & \makebox[0.266\columnwidth][c]{\scriptsize\textbf{GT}}%
    \makebox[0.266\columnwidth][c]{\scriptsize\textbf{MediaPipe Temporal}}%
    \makebox[0.266\columnwidth][c]{\scriptsize\textbf{Ours}} \\[1.5mm]

  \centering\raisebox{0.8em}{\scriptsize\textbf{In-frame}} & \includegraphics[width=0.80\columnwidth]{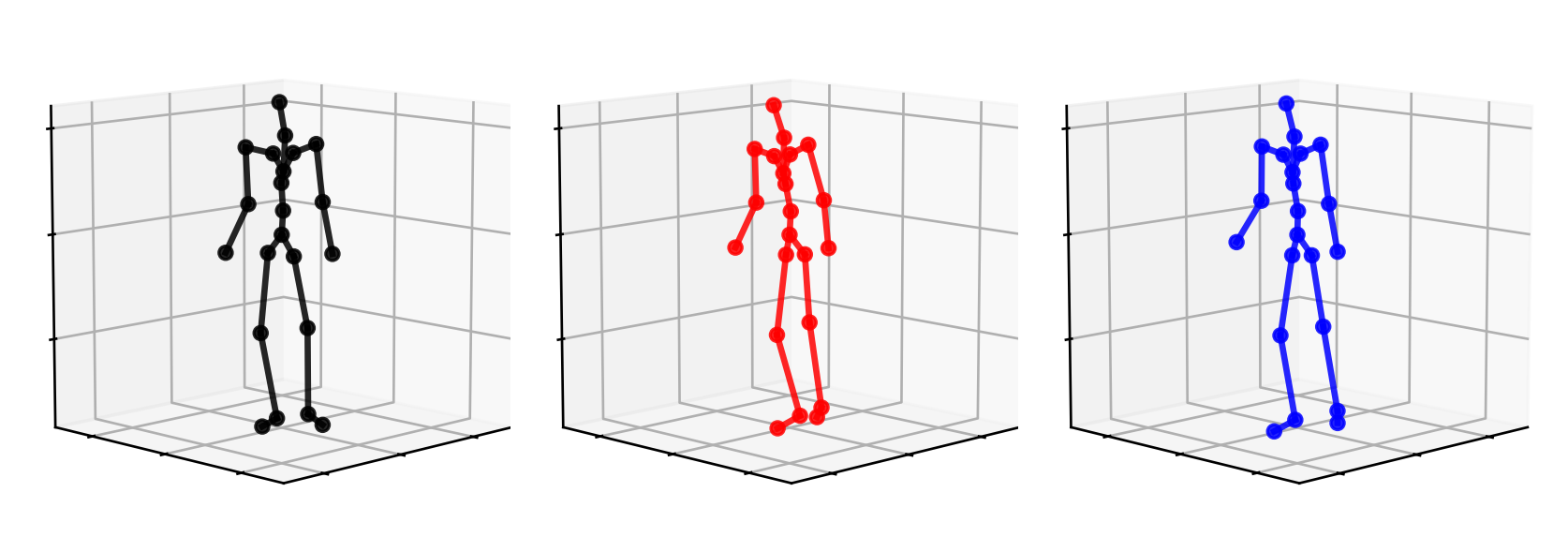} \\[1.5mm]
  
  \centering\raisebox{0.8em}{\scriptsize\textbf{Dropout}}  & \includegraphics[width=0.80\columnwidth]{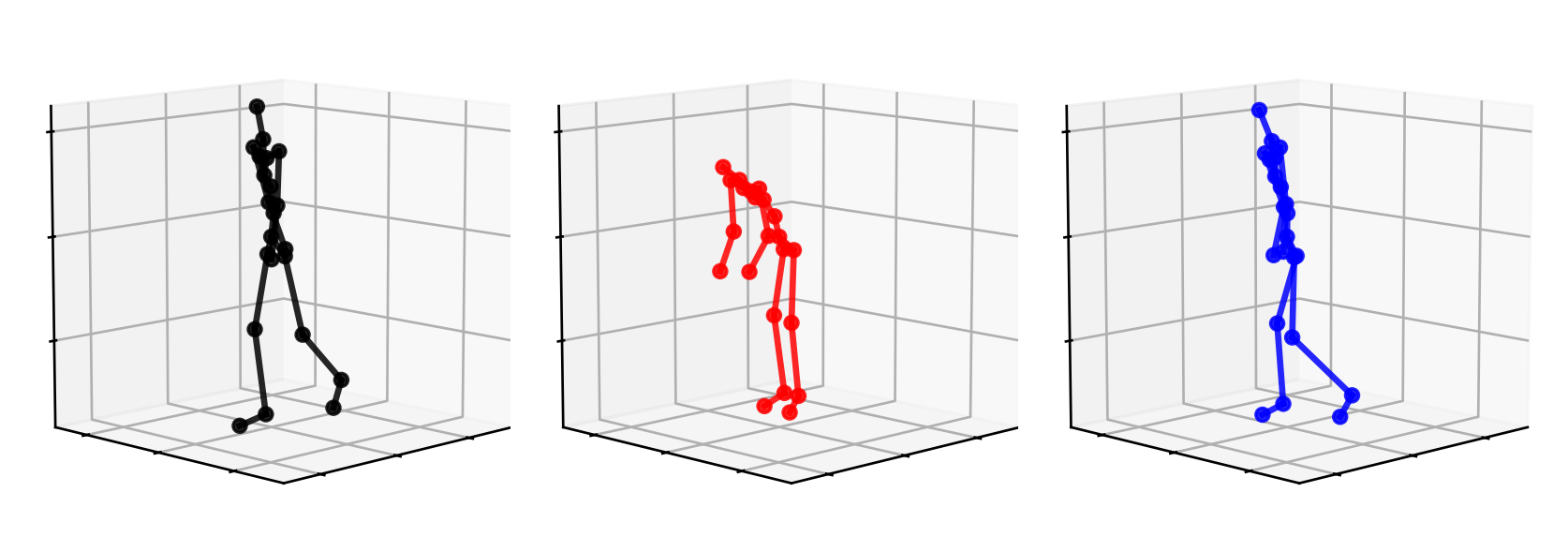} \\[1.5mm]
  
  \centering\raisebox{0.8em}{\scriptsize\textbf{Dropout}}  & \includegraphics[width=0.80\columnwidth]{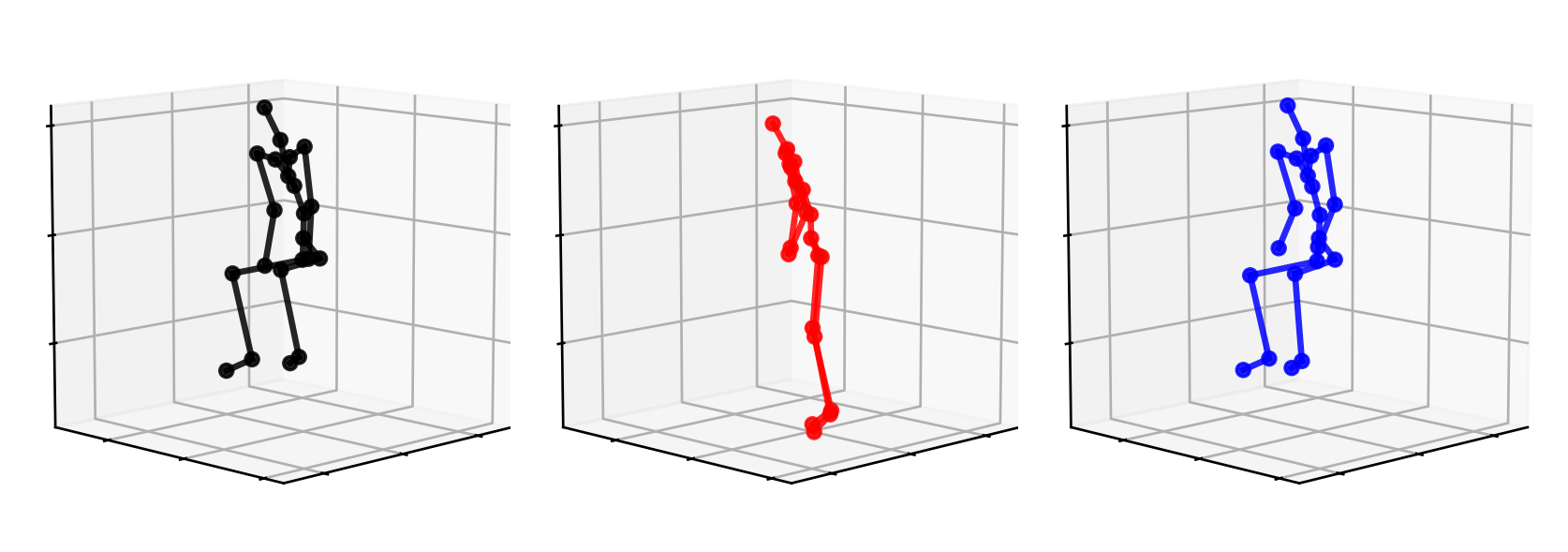}
\end{tabular}
\caption{Qualitative behavior under short visual dropout. Rows show three temporal frames in order: an in-frame moment followed by two dropout moments. In each row, the columns from left to right are GT, MediaPipe Temporal, and Ours. During dropout, the vision-only baseline has no new visual observation and therefore degenerates to last-observation holding or temporal-prior extrapolation. The proposed model continues to update the articulated pose using pressure and acceleration cues from the smart insole.}
\label{fig:out_of_frame}
\end{figure}

\subsection{Ablation Study}

\begin{table}[t]
\centering
\caption{Ablation study on the MediaPipe clean setting. All metrics are lower-is-better.}
\label{tab:ablation}
\renewcommand{\arraystretch}{1.35} 
\setlength{\tabcolsep}{3pt} 
\resizebox{\columnwidth}{!}{
\begin{tabular}{lccc}
\toprule
Method & MPJPE (mm) $\downarrow$ & PA-MPJPE (mm) $\downarrow$ & Jitter ($\mathrm{m/s^3}$) $\downarrow$ \\
\midrule
Vision Temporal      & 67.55 & 45.21 & 81.40 \\
Naive Temporal Fusion& 66.90 & 43.09 & 77.05 \\
w/o Coordinator      & 66.25 & 43.88 & 80.25 \\
w/o Pressure         & 65.07 & 42.20 & 73.00 \\
w/o Acceleration     & 66.73 & 43.55 & 67.38 \\
Ours                 & \textbf{64.15} & \textbf{40.62} & \textbf{65.05} \\
\bottomrule
\end{tabular}
}
\end{table}

Table~\ref{tab:ablation} presents the ablation results on the MediaPipe clean setting. The vision-only temporal baseline already provides a strong comparison because it uses temporal modeling without insole input. Directly concatenating visual and insole features, denoted as Naive Temporal Fusion, improves MPJPE and PA-MPJPE over the vision-only baseline, showing that foot sensing is useful even with a simple fusion design. However, it remains inferior to the proposed part-decoupled fusion network.

Removing the global coordinator leads to worse PA-MPJPE and jitter, indicating that cross-branch information exchange is needed to maintain whole-body consistency after separating upper- and lower-body streams. Removing pressure or acceleration also degrades the performance. Without pressure, the model loses direct contact and load-transfer cues; without acceleration, it loses dynamic information during swing phases. The full model achieves the best performance across all three metrics, demonstrating that both the sensing modalities and the part-decoupled fusion design contribute to the final improvement.

The two sensing components play complementary roles. Plantar pressure is more directly related to stance, support phase, and load transfer, so it is especially useful for correcting foot-contact timing and lower-body geometry when visual evidence is weak. Foot-mounted acceleration provides dynamic information during swing phases, where pressure may be close to zero and contact cues alone become ambiguous. This explains why removing either modality degrades the result, and why the full pressure--inertial representation provides the most consistent improvement.

\subsection{Short Visual-Dropout Analysis}

In the previous sections (Tables~\ref{tab:main_results} and \ref{tab:gait_results}), we quantitatively demonstrated the exceptional robustness of the proposed method under short visual-dropout conditions. To intuitively understand the underlying tracking mechanism during these extreme visual degradation scenarios, we conduct a qualitative behavioral analysis.

This setting is fundamentally different from lower-body occlusion. Under lower-body occlusion, the visual backbone still receives partial image evidence (e.g., upper body or surrounding context). During out-of-frame dropout, however, the visual stream provides absolutely no valid observation of the ongoing motion. In our implementation, the vision-only baseline therefore completely freezes, degenerating to last-observation holding or unpredictable temporal-prior extrapolation. 

As illustrated in Figure~\ref{fig:out_of_frame}, the rows show continuous temporal frames transitioning from an in-frame moment to an out-of-frame dropout. When the visual stream becomes unavailable, the vision-only MediaPipe baseline remains stuck in its last visible posture. However, our proposed model continues to accurately update lower-body articulation (e.g., knee lifting and step advancing) using continuous plantar pressure and foot-mounted acceleration. This result visually corroborates the quantitative metrics: while foot sensing is not intended to replace vision for infinite long-duration blind tracking, it decisively provides the essential physical observability needed to survive short, inevitable camera failures in real-world deployments.

\subsection{Mobile Runtime Overhead}

We further benchmark the runtime overhead of the proposed fusion module on three consumer Android smartphones. Since the proposed method is designed to operate after a real-time visual pose backbone has produced pose features, this benchmark focuses on the additional neural inference latency introduced by the proposed vision--pressure--inertial fusion module. We use pre-recorded synchronized and normalized visual and foot-sensing feature streams as frame-by-frame inputs, and run the model in streaming mode with batch size 1 while carrying recurrent hidden states online. Data loading and 200 warm-up frames are excluded from the latency measurement.

As shown in Table~\ref{tab:runtime}, the proposed fusion module achieves sub-1.5 ms average latency and sub-1.5 ms p95 latency on all tested devices, including an older mid-range phone. Even on the OPPO Reno3 5G, the p95 latency is only 1.47 ms, far below the 33.33 ms frame budget required for 30 FPS operation. These results indicate that the proposed fusion module introduces negligible computational overhead once real-time visual pose features are available. The model size is 12.10 MB in ONNX format.

It should be noted that this runtime measures only the proposed fusion module after visual pose features have been produced. The end-to-end system latency depends on the selected visual backbone, camera pipeline, and platform-specific preprocessing. This design is intentional: the proposed module is not tied to a particular visual estimator, and its role is to provide a lightweight pressure--inertial refinement layer that can be attached to different real-time pose backbones.

\begin{table}[t]
\centering
\caption{Mobile runtime overhead of the proposed fusion module. Latency is measured using ONNX Runtime Android on CPU with synchronized feature streams, batch size 1, streaming inference, and 200 warm-up frames. The ONNX model size is 12.10 MB.}
\label{tab:runtime}
\setlength{\tabcolsep}{3.5pt}
\resizebox{\columnwidth}{!}{
\begin{tabular}{llcccc}
\toprule
Device & SoC / Chipset (Node) & Avg. & p95 & p99 & FPS \\
& & (ms) & (ms) & (ms) & \\
\midrule
vivo X200 Pro & MediaTek Dimensity 9400 (3 nm) & 0.44 & 0.68 & 0.80 & 2264.46 \\
Redmi Note 12 & Snapdragon 7+ Gen 2 (4 nm) & 1.04 & 1.15 & 1.33 & 960.88 \\
OPPO Reno3 & MediaTek Dimensity 1000L (7 nm) & 1.32 & 1.47 & 2.00 & 758.59 \\
\bottomrule
\end{tabular}
}
\end{table}

\section{Conclusion and Future Work}
\label{sec:conclusion}

This paper presented a part-decoupled vision--pressure--inertial fusion framework for real-time monocular human pose estimation. We investigated whether sparse plantar-pressure and foot-acceleration signals can provide complementary cues when monocular visual evidence is incomplete or unavailable. The framework maps heterogeneous visual observations into a unified root-relative skeleton space, incorporates foot sensing into lower-body temporal modeling, and uses a lightweight global coordinator to maintain whole-body consistency.

Experimental results show that the insole-enhanced framework improves pose accuracy and temporal smoothness over vision-only temporal baselines under clean conditions, with larger gains under lower-body occlusion. Analysis of short visual dropouts further demonstrates the value of continuous foot-contact and foot-dynamic cues. The method also improves contact-sensitive gait measurements, including stance accuracy, step-distance error, and cadence error. These results indicate that sparse pressure--inertial foot sensing provides useful physical priors for robust human motion estimation.

Although the experiments use I-SOL smart insoles, the formulation only assumes synchronized sparse plantar-pressure and foot-mounted acceleration streams with comparable temporal resolution. The method is therefore not limited to a specific insole model. Data organization, action definitions, and preprocessing scripts will be released to support reproducibility.

The method nevertheless has several limitations. First, it estimates root-relative pose rather than long-term global trajectories. Because the current sensing setup lacks reliable global orientation, extended visual dropouts may still cause drift. Second, broader evaluation across users, environments, and motion types is needed. Third, downstream applications such as ergonomic risk assessment, rehabilitation scoring, and abnormal gait screening require further validation.

Future work will improve long-horizon robustness through lightweight global motion constraints and explicit contact-state modeling. We will also explore mobile and edge deployment with real-time visual front-ends and Bluetooth insole streams, and extend evaluation to industrial monitoring, home rehabilitation, and long-term daily activity assessment.

\bibliographystyle{IEEEtran} 
\bibliography{myrefs}    

@InProceedings{Maji_2022_CVPR,
    author    = {Maji, Debapriya and Nagori, Soyeb and Mathew, Manu and Poddar, Deepak},
    title     = {YOLO-Pose: Enhancing YOLO for Multi Person Pose Estimation Using Object Keypoint Similarity Loss},
    booktitle = {Proceedings of the IEEE/CVF Conference on Computer Vision and Pattern Recognition (CVPR) Workshops},
    month     = {June},
    year      = {2022},
    pages     = {2637--2646}
}

@article{bazarevsky2020blazepose,
  title={BlazePose: On-device Real-time Body Pose Tracking},
  author={Bazarevsky, Valentin and Grishchenko, Ivan and Raveendran, Karthik and Zhu, Tyler and Zhang, Fan and Grundmann, Matthias},
  journal={arXiv preprint arXiv:2006.10204},
  year={2020}
}

@InProceedings{Su_2025_CVPR,
    author    = {Su, Chi and Ma, Xiaoxuan and Su, Jiajun and Wang, Yizhou},
    title     = {SAT-HMR: Real-Time Multi-Person 3D Mesh Estimation via Scale-Adaptive Tokens},
    booktitle = {Proceedings of the IEEE/CVF Conference on Computer Vision and Pattern Recognition (CVPR)},
    month     = {June},
    year      = {2025},
    pages     = {16796--16806}
}

@ARTICLE{cai2024playing,
  author={Cai, Zhongang and Zhang, Mingyuan and Ren, Jiawei and Wei, Chen and Ren, Daxuan and Lin, Zhengyu and Zhao, Haiyu and Yang, Lei and Loy, Chen Change and Liu, Ziwei},
  journal={IEEE Transactions on Pattern Analysis and Machine Intelligence}, 
  title={Playing for 3D Human Recovery}, 
  year={2024},
  volume={46},
  number={12},
  pages={10533--10545},
  doi={10.1109/TPAMI.2024.3450537}
}

@inproceedings{shin2024wham,
  title={WHAM: Reconstructing World-Grounded Humans with Accurate 3D Motion},
  author={Shin, Soyong and Kim, Juyong and Halilaj, Eni and Black, Michael J.},
  booktitle={Proceedings of the IEEE/CVF Conference on Computer Vision and Pattern Recognition (CVPR)},
  month={June},
  pages={2070--2080},
  year={2024}
}

@inproceedings{wang2025prompthmr,
  title={PromptHMR: Promptable Human Mesh Recovery},
  author={Wang, Yufu and Sun, Yu and Patel, Priyanka and Daniilidis, Kostas and Black, Michael J. and Kocabas, Muhammed},
  booktitle={Proceedings of the IEEE/CVF Conference on Computer Vision and Pattern Recognition (CVPR)},
  pages={1148--1159},
  year={2025}
}

@ARTICLE{peng2024plantar,
  author={Peng, Sai and Hassan, Hytham and Rosseel, Stijn and Matricali, Giovanni A. and Deschamps, Kevin and Vandeginste, Veerle and Hallez, Hans},
  journal={IEEE Sensors Journal}, 
  title={Recent Advances in 3-D Printed, Wearable Pressure Sensors for Plantar Pressure Monitoring: A Review}, 
  year={2024},
  volume={24},
  number={21},
  pages={33903--33921},
  doi={10.1109/JSEN.2024.3457040}
}

@INPROCEEDINGS{weigand2022imu,
  author={Weigand, Florian and Höhl, Andreas and Zeiss, Julian and Konigorski, Ulrich and Grimmer, Martin},
  booktitle={2022 IEEE/RSJ International Conference on Intelligent Robots and Systems (IROS)}, 
  title={Continuous Locomotion Mode Recognition and Gait Phase Estimation Based on a Shank-Mounted IMU with Artificial Neural Networks}, 
  year={2022},
  pages={12744--12751},
  doi={10.1109/IROS47612.2022.9981154}
}

@inproceedings{wu2024soleposer,
  title={SolePoser: Full Body Pose Estimation Using a Single Pair of Insole Sensor},
  author={Wu, Erwin and Khirodkar, Rawal and Koike, Hideki and Kitani, Kris},
  booktitle={Proceedings of the 37th Annual ACM Symposium on User Interface Software and Technology (UIST)},
  pages={67:1--67:9},
  year={2024},
  doi={10.1145/3654777.3676418}
}

@article{ying2025form,
  author    = {Ying, Qijun and Cao, Zehua and Wu, Ziyu and Deng, Wenwu and Zhong, Yuchen and Diao, Yukun and Cai, XiaoHui},
  title     = {FoRM: Foot-Driven Reconstruction of Human Motion Using Dual-Modal Plantar Pressure and Inertial Sensing},
  journal   = {Proceedings of the ACM on Interactive, Mobile, Wearable and Ubiquitous Technologies},
  volume    = {9},
  number    = {3},
  articleno = {145},
  year      = {2025},
  doi       = {10.1145/3749551}
}

@InProceedings{tripathi2023physics,
    author    = {Tripathi, Shashank and M\"uller, Lea and Huang, Chun-Hao P. and Taheri, Omid and Black, Michael J. and Tzionas, Dimitrios},
    title     = {3D Human Pose Estimation via Intuitive Physics},
    booktitle = {Proceedings of the IEEE/CVF Conference on Computer Vision and Pattern Recognition (CVPR)},
    month     = {June},
    year      = {2023},
    pages     = {4713--4725}
}

@inproceedings{zhang2024mmvp,
  title={MMVP: A Multimodal MoCap Dataset with Vision and Pressure Sensors},
  author={Zhang, He and Ren, Shenghao and Yuan, Haolei and Zhao, Jianhui and Li, Fan and Sun, Shuangpeng and Liang, Zhenghao and Yu, Tao and Shen, Qiu and Cao, Xun},
  booktitle={Proceedings of the IEEE/CVF Conference on Computer Vision and Pattern Recognition (CVPR)},
  month={June},
  pages={21842--21852},
  year={2024}
}

@InProceedings{ren2025motionpro,
    author    = {Ren, Shenghao and Lu, Yi and Huang, Jiayi and Zhao, Jiayi and Zhang, He and Yu, Tao and Shen, Qiu and Cao, Xun},
    title     = {MotionPRO: Exploring the Role of Pressure in Human MoCap and Beyond},
    booktitle = {Proceedings of the IEEE/CVF Conference on Computer Vision and Pattern Recognition (CVPR)},
    month     = {June},
    year      = {2025},
    pages     = {27760--27770}
}

@article{fang2023alphapose,
  title={AlphaPose: Whole-Body Regional Multi-Person Pose Estimation and Tracking in Real-Time},
  author={Fang, Hao-Shu and Li, Jiefeng and Tang, Hongyang and Xu, Chao and Zhu, Haoyi and Xiu, Yuliang and Li, Yong-Lu and Lu, Cewu},
  journal={IEEE Transactions on Pattern Analysis and Machine Intelligence},
  volume={45},
  number={6},
  pages={7157--7173},
  year={2023},
  doi={10.1109/TPAMI.2022.3222784}
}

@inproceedings{lu2024rtmo,
  title={RTMO: Towards High-Performance One-Stage Real-Time Multi-Person Pose Estimation},
  author={Lu, Peng and Jiang, Tao and Li, Yining and Li, Xiangtai and Chen, Kai and Yang, Wenming},
  booktitle={Proceedings of the IEEE/CVF Conference on Computer Vision and Pattern Recognition (CVPR)},
  pages={1491--1500},
  year={2024}
}

@inproceedings{zhao2023poseformerv2,
  title={PoseFormerV2: Exploring Frequency Domain for Efficient and Robust 3D Human Pose Estimation},
  author={Zhao, Qitao and Zheng, Ce and Liu, Mengyuan and Wang, Pichao and Chen, Chen},
  booktitle={Proceedings of the IEEE/CVF Conference on Computer Vision and Pattern Recognition (CVPR)},
  year={2023}
}

@inproceedings{zhu2023motionbert,
  title={MotionBERT: A Unified Perspective on Learning Human Motion Representations},
  author={Zhu, Wentao and Ma, Xiaoxuan and Liu, Zhaoyang and Liu, Libin and Wu, Wayne and Wang, Yizhou},
  booktitle={Proceedings of the IEEE/CVF International Conference on Computer Vision (ICCV)},
  year={2023}
}

@inproceedings{goel2023humans,
  title={Humans in 4D: Reconstructing and Tracking Humans with Transformers},
  author={Goel, Shubham and Pavlakos, Georgios and Rajasegaran, Jathushan and Kanazawa, Angjoo and Malik, Jitendra},
  booktitle={Proceedings of the IEEE/CVF International Conference on Computer Vision (ICCV)},
  year={2023}
}

@inproceedings{gong2023diffpose,
  title={DiffPose: Toward More Reliable 3D Pose Estimation},
  author={Gong, Jia and Foo, Lin Geng and Fan, Zhipeng and Ke, Qiuhong and Rahmani, Hossein and Liu, Jun},
  booktitle={Proceedings of the IEEE/CVF Conference on Computer Vision and Pattern Recognition (CVPR)},
  pages={13041--13051},
  year={2023}
}

@inproceedings{sun2024repose,
  title={RePOSE: 3D Human Pose Estimation via Spatio-Temporal Depth Relational Consistency},
  author={Sun, Ziming and Liang, Yuan and Ma, Zejun and Zhang, Tianle and Bao, Linchao and Li, Guiqing and He, Shengfeng},
  booktitle={European Conference on Computer Vision (ECCV)},
  pages={309--325},
  year={2024}
}

@article{gao2024pressinpose,
  title={PressInPose: Integrating Pressure and Inertial Sensors for Full-Body Pose Estimation in Activities},
  author={Gao, Yang and Zhang, Wenbo and Ren, Junbin and Zheng, Ruihao and Jin, Yingcheng and Wu, Di and Shu, Lin and Xu, Xiangmin and Jin, Zhanpeng},
  journal={Proceedings of the ACM on Interactive, Mobile, Wearable and Ubiquitous Technologies},
  volume={8},
  number={4},
  pages={197:1--197:28},
  year={2024},
  doi={10.1145/3699773}
}

@inproceedings{zhang2024dynaip,
  title={Dynamic Inertial Poser (DynaIP): Part-Based Motion Dynamics Learning for Enhanced Human Pose Estimation with Sparse Inertial Sensors},
  author={Zhang, Yu and Xia, Songpengcheng and Chu, Lei and Yang, Jiarui and Wu, Qi and Pei, Ling},
  booktitle={Proceedings of the IEEE/CVF Conference on Computer Vision and Pattern Recognition (CVPR)},
  pages={1889--1899},
  year={2024}
}

@ARTICLE{xu2026opensim,
  author={Xu, Chenquan and Tan, Yuanshuo and Strout, Zach and Liu, Guoxing and Zhu, Kezhe and Wang, Hong and Shull, Peter B.},
  journal={IEEE Transactions on Neural Systems and Rehabilitation Engineering}, 
  title={Real-Time OpenSim via IMUs for Full Body Kinematics During Gait, Sports, Exercise, and Dance Movements}, 
  year={2026},
  volume={34},
  number={},
  pages={650-662},
  doi={10.1109/TNSRE.2026.3653477}}

@article{jiang2023rtmpose,
  title={RTMPose: Real-Time Multi-Person Pose Estimation Based on MMPose},
  author={Jiang, Tao and Lu, Peng and Zhang, Li and Ma, Ningsheng and Han, Rui and Lyu, Chengqi and Li, Yining and Chen, Kai},
  journal={arXiv preprint arXiv:2303.07399},
  year={2023}
}

@inproceedings{yang2023dwpose,
  title={Effective Whole-Body Pose Estimation with Two-Stages Distillation},
  author={Yang, Zhendong and Zeng, Ailing and Yuan, Chun and Li, Yu},
  booktitle={Proceedings of the IEEE/CVF International Conference on Computer Vision Workshops},
  pages={4210--4220},
  year={2023}
}

@inproceedings{xu2022vitpose,
  title={ViTPose: Simple Vision Transformer Baselines for Human Pose Estimation},
  author={Xu, Yufei and Zhang, Jing and Zhang, Qiming and Tao, Dacheng},
  booktitle={Advances in Neural Information Processing Systems},
  volume={35},
  pages={38571--38584},
  year={2022}
}

@inproceedings{mehraban2024motionagformer,
  title={MotionAGFormer: Enhancing 3D Human Pose Estimation with a Transformer-GCNFormer Network},
  author={Mehraban, Soroush and Adeli, Vida and Taati, Babak},
  booktitle={Proceedings of the IEEE/CVF Winter Conference on Applications of Computer Vision},
  pages={6920--6930},
  year={2024}
}

@inproceedings{xu2024mobileposer,
  title={MobilePoser: Real-Time Full-Body Pose Estimation and 3D Human Translation from IMUs in Mobile Consumer Devices},
  author={Xu, Vasco and Gao, Chenfeng and Hoffmann, Henry and Ahuja, Karan},
  booktitle={Proceedings of the 37th Annual ACM Symposium on User Interface Software and Technology (UIST)},
  articleno={70},
  pages={70:1--70:11},
  year={2024},
  doi={10.1145/3654777.3676461}
}

@inproceedings{pavllo20193d,
  title={3d human pose estimation in video with temporal convolutions and semi-supervised training},
  author={Pavllo, Dario and Feichtenhofer, Christoph and Grangier, David and Auli, Michael},
  booktitle={Proceedings of the IEEE/CVF conference on computer vision and pattern recognition},
  pages={7753--7762},
  year={2019}
}

@inproceedings{kocabas2020vibe,
  title={Vibe: Video inference for human body pose and shape estimation},
  author={Kocabas, Muhammed and Athanasiou, Nikos and Black, Michael J},
  booktitle={Proceedings of the IEEE/CVF conference on computer vision and pattern recognition},
  pages={5253--5263},
  year={2020}
}

@inproceedings{zhou2017towards,
  title={Towards 3d human pose estimation in the wild: a weakly-supervised approach},
  author={Zhou, Xingyi and Huang, Qixing and Sun, Xiao and Xue, Xiangyang and Wei, Yichen},
  booktitle={Proceedings of the IEEE international conference on computer vision},
  pages={398--407},
  year={2017}
}

@ARTICLE{li2025p2lhap,
  author={Li, Shuangjian and Zhu, Tao and Nie, Mingxing and Ning, Huansheng and Liu, Zhenyu and Chen, Liming},
  journal={IEEE Internet of Things Journal}, 
  title={P2LHAP: Wearable-Sensor-Based Human Activity Recognition, Segmentation, and Forecast Through Patch-to-Label Seq2Seq Transformer}, 
  year={2025},
  volume={12},
  number={6},
  pages={6818-6830},
  doi={10.1109/JIOT.2024.3493380}}

@ARTICLE{li2025harmamba,
  author={Li, Shuangjian and Zhu, Tao and Duan, Furong and Chen, Liming and Ning, Huansheng and Nugent, Christopher and Wan, Yaping},
  journal={IEEE Internet of Things Journal}, 
  title={HARMamba: Efficient and Lightweight Wearable Sensor Human Activity Recognition Based on Bidirectional Mamba}, 
  year={2025},
  volume={12},
  number={3},
  pages={2373-2384},
  doi={10.1109/JIOT.2024.3463405}}

@ARTICLE{fang2024spgformer,
  author={Fang, Qin and Xu, Zihan and Hu, Mengxian and Zeng, Qinyang and Liu, Chengju and Chen, Qijun},
  journal={IEEE Transactions on Instrumentation and Measurement},
  title={{SPGformer}: Serial--Parallel Hybrid {GCN}-Transformer With Graph-Oriented Encoder for 2-D-to-3-D Human Pose Estimation},
  year={2024},
  volume={73},
  pages={1--15},
  doi={10.1109/TIM.2024.3381701}
}

\begin{IEEEbiography} [{\includegraphics[width=1in,height=1.25in,clip,keepaspectratio] {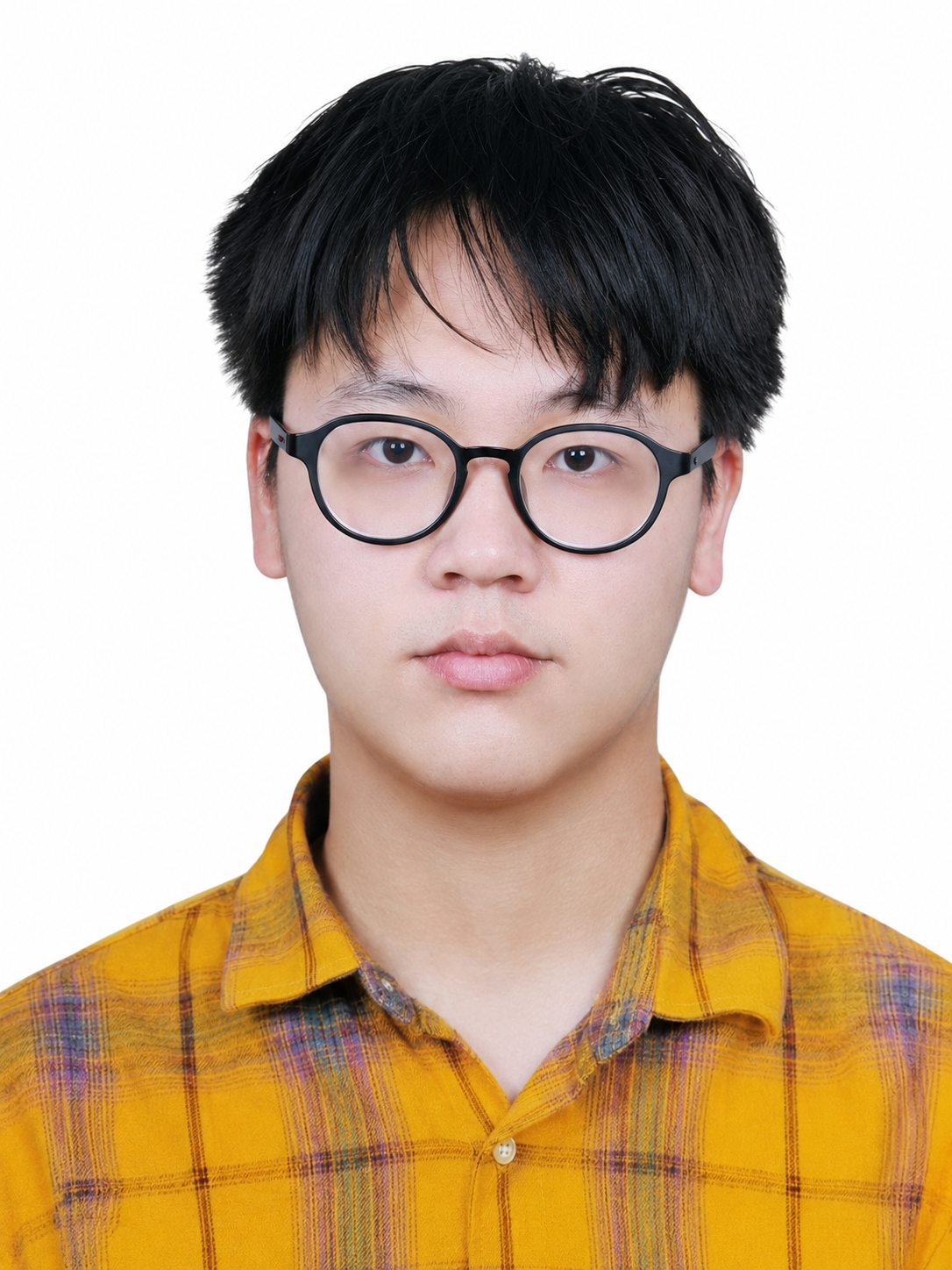}}] {Liang Yu} received the bachelor's degree from Qinghai Minzu University, Xining, China. He is currently pursuing the M.S. degree with the School of Computer Science, University of South China, Hengyang, China. His research interests include intelligent perception and pattern recognition. \end{IEEEbiography}

\begin{IEEEbiography} [{\includegraphics[width=1in,height=1.25in,clip,keepaspectratio] {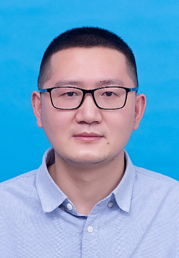}}] {Tao Zhu} (Senior Member, IEEE) received the B.E. degree from Central South University, Changsha, China, in 2009, and the Ph.D. degree from the University of Science and Technology of China, Hefei, China, in 2015. He is currently an Associate Professor with the University of South China, Hengyang, China. He has served as the Principal Investigator of several projects funded by the National Natural Science Foundation of China and the Natural Science Foundation of Hunan Province. His research interests include the Internet of Things, pervasive computing, assisted living, and evolutionary computation. Dr. Zhu is currently the Chair of the IEEE Computational Intelligence Society Smart World Technical Committee Task Force on User-Centred Smart Systems. \end{IEEEbiography}

\begin{IEEEbiography} [{\includegraphics[width=1in,height=1.25in,clip,keepaspectratio] {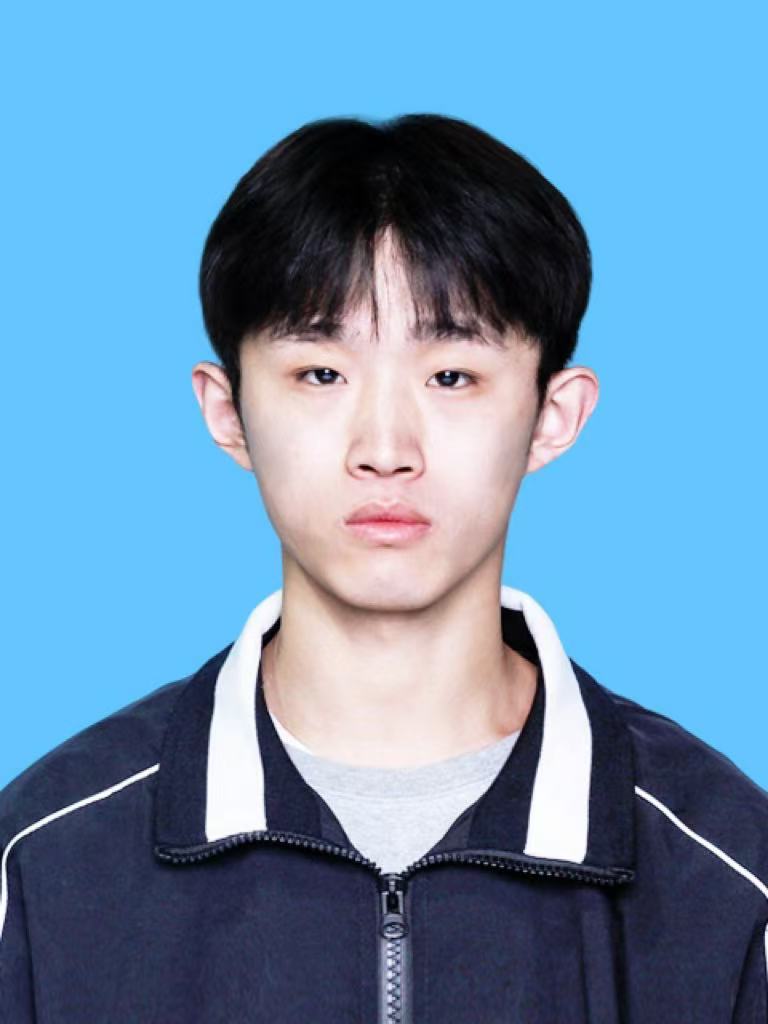}}] {Xiangqing Lu} received the bachelor's degree from Qilu University of Technology (Shandong Academy of Sciences), Jinan, China. He is currently pursuing the M.S. degree with the School of Computer Science, University of South China, Hengyang, China. His research interests include intelligent perception and pattern recognition. \end{IEEEbiography}

\begin{IEEEbiography} [{\includegraphics[width=1in,height=1.25in,clip,keepaspectratio] {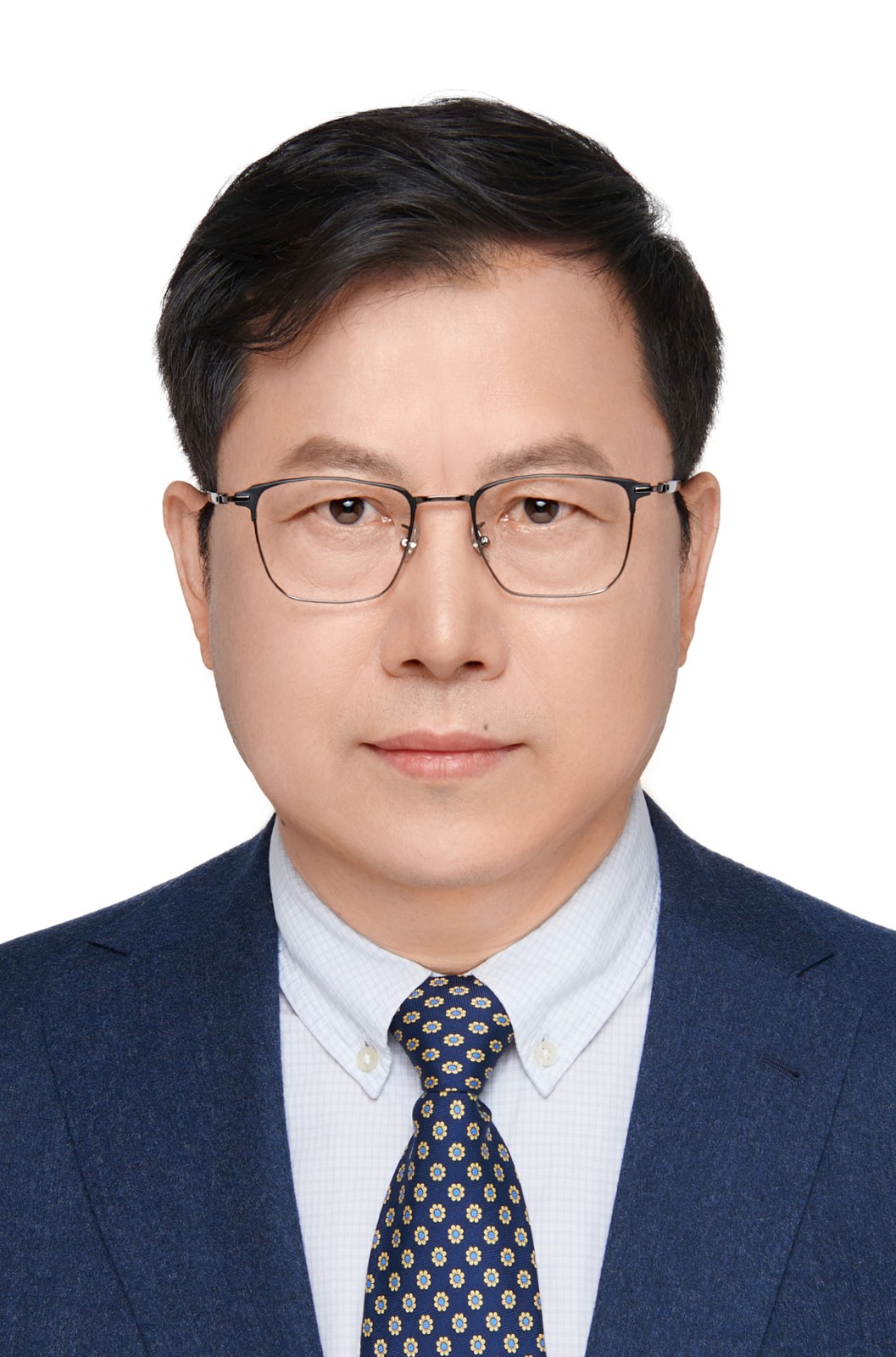}}] {Liming Chen} (Senior Member, IEEE) received the M.S. degree in vehicle engineering from the Beijing Institute of Technology, Beijing, China, in 1988, and the Ph.D. degree in computer science from De Montfort University, Leicester, U.K., in 2001. He is currently a Professor with the School of Computer Science and Technology, Dalian University of Technology, Dalian, China. His research interests include pervasive computing, data analytics, artificial intelligence, user-centered intelligent systems, digital healthcare, and cybersecurity. He has published more than 250 papers in these areas. Prof. Chen is a Fellow of the Institution of Engineering and Technology. \end{IEEEbiography}

\begin{IEEEbiography} [{\includegraphics[width=1in,height=1.25in,clip,keepaspectratio] {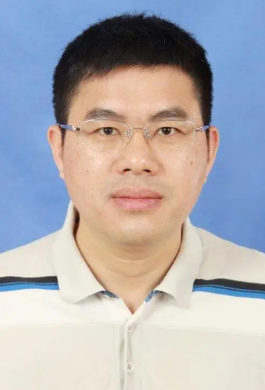}}] {Wenjian Luo} (Senior Member, IEEE) received the B.S. degree in computer software and the Ph.D. degree in computer application technology from the University of Science and Technology of China, Hefei, China, in 1998 and 2003, respectively. He is currently a Professor with the School of Computer Science and Technology, Harbin Institute of Technology, Shenzhen, China. His research interests include artificial intelligence security, privacy-preserving computing, big-data analytics, machine learning, immune computing, and swarm intelligence. He is a Distinguished Member of CCF and a Senior Member of ACM and CAAI. \end{IEEEbiography}

\vfill %

\end{document}